\pdfoutput=1

\documentclass[11pt]{article}

\usepackage{acl}

\usepackage{times}
\usepackage{latexsym}

\usepackage[T1]{fontenc}

\usepackage[utf8]{inputenc}

\usepackage{microtype}

\usepackage{inconsolata}

\usepackage{amsmath}
\usepackage{amsthm}
\usepackage{amsfonts}
\usepackage{amssymb}
\usepackage{tikz}
\usepackage{cleveref}
\usepackage{booktabs}
\usepackage{bbm}
\usepackage{mathtools}
\usepackage{multirow}

\usepackage{soul}
\newcommand{\highlight}[2]{%
    \begingroup
    \definecolor{hlcolor}{HTML}{#1}%
    \sethlcolor{hlcolor}%
    \hl{#2}%
    \endgroup
}
\newcommand{\token}[1]{\highlight{DFDFDF}{#1}}
\usepackage{subcaption}
\usepackage{adjustbox}
\usepackage{examples-slim}
\newcommand{\eg}[1]{(\ref{#1})}

\theoremstyle{definition}

\newcommand{\feature}[1]{\mathbf{a}_{\text{#1}}}
\newcommand{\pair}[2]{[\textbf{#1}/\textbf{#2}]}
\newcommand{\benchmarktitle}{\textbf{\texttt{CausalGym}}}

\newcommand{\intlm}{p_{f \gets f^*}}
\newcommand{\diilm}{p_{f \gets f^*_{\mathbf{a}}}}
\newcommand{\diilmf}[1]{p_{f \gets f^*_{\feature{#1}}}}

\crefformat{section}{\S#2#1#3}
\crefformat{subsection}{\S#2#1#3}
\crefformat{subsubsection}{\S#2#1#3}
\crefformat{paragraph}{\P#2#1#3}
\crefformat{subparagraph}{\P#2#1#3}
\crefmultiformat{section}{\S#2#1#3}{ and~\S#2#1#3}{, \S#2#1#3}{, and~\S#2#1#3}
\crefmultiformat{subsection}{\S#2#1#3}{ and~\S#2#1#3}{, \S#2#1#3}{, and~\S#2#1#3}
\crefmultiformat{subsubsection}{\S#2#1#3}{ and~\S#2#1#3}{, \S#2#1#3}{, and~\S#2#1#3}
\crefmultiformat{paragraph}{\P\P#2#1#3}{ and~#2#1#3}{, #2#1#3}{, and~#2#1#3}
\crefmultiformat{subparagraph}{\P\P#2#1#3}{ and~#2#1#3}{, #2#1#3}{, and~#2#1#3}
\crefrangeformat{section}{\mbox{\S\S#3#1#4--#5#2#6}}
\crefrangeformat{subsection}{\mbox{\S\S#3#1#4--#5#2#6}}
\crefrangeformat{subsubsection}{\mbox{\S\S#3#1#4--#5#2#6}}
\crefrangeformat{paragraph}{\mbox{\P\P#3#1#4--#5#2#6}}
\crefrangeformat{subparagraph}{\mbox{\P\P#3#1#4--#5#2#6}}
\crefname{part}{Part}{Parts}
\Crefname{part}{Part}{Parts}
\crefname{chapter}{Ch.}{Ch.}
\Crefname{chapter}{Ch.}{Ch.}
\crefname{footnote}{Fn.}{Fn.}
\Crefname{footnote}{Fn.}{Fn.}
\crefname{figure}{Figure}{Figures}
\crefname{table}{Table}{Tables}
\crefname{subfigure}{Figure}{Figures}
\Crefname{subfigure}{Figure}{Figures}
\crefname{appsec}{Appendix}{Appendices}
\Crefname{appsec}{Appendix}{Appendices}
\crefname{algocf}{Algorithm}{Algorithms}
\Crefname{algocf}{Algorithm}{Algorithms}
\crefname{xnumi}{ex.}{exs.}
\Crefname{xnumi}{Ex.}{Exs.}
\crefname{xnumii}{ex.}{exs.}
\Crefname{xnumii}{Ex.}{Exs.}

\title{\benchmarktitle{}: Benchmarking causal interpretability methods \\ on linguistic tasks}

\author{Aryaman Arora~\;~Dan Jurafsky~\;~Christopher Potts \\
  Stanford University \\
  \texttt{\{aryamana,jurafsky,cgpotts\}@stanford.edu}}

\begin{document}
\maketitle
\begin{abstract}
Language models (LMs) have proven to be powerful tools for psycholinguistic research, but most prior work has focused on purely behavioural measures (e.g., surprisal comparisons). At the same time, research in model interpretability has begun to illuminate the abstract causal mechanisms shaping LM behavior. To help bring these strands of research closer together, we introduce \benchmarktitle{}. We adapt and expand the SyntaxGym suite of tasks to benchmark the ability of interpretability methods to causally affect model behaviour. To illustrate how \benchmarktitle{} can be used, we study the \texttt{pythia} models (14M--6.9B) and assess the causal efficacy of a wide range of interpretability methods, including linear probing and distributed alignment search (DAS). 
We find that DAS outperforms the other methods, and so we use it to study the learning trajectory of two difficult linguistic phenomena in \texttt{pythia-1b}: negative polarity item licensing and filler--gap dependencies. Our analysis shows that the mechanism implementing both of these tasks is learned in discrete stages, not gradually.
\newline
\newline
\hspace{.5em}\includegraphics[width=1.25em,height=1.25em]{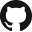}\hspace{.75em}\parbox{\dimexpr\linewidth-2\fboxsep-2\fboxrule}{\url{https://github.com/aryamanarora/causalgym}}
\end{abstract}

\section{Introduction}
\label{sec:intro}

Language models have found increasing use as tools for psycholinguistic investigation---to model word surprisal \citep[][\textit{inter alia}]{smith2013effect,goodkind-bicknell-2018-predictive,wilcox-etal-2023-language,shainetal24}, graded grammaticality judgements \citep{hu2024language}, and, broadly, human language processing \citep{futrell-etal-2019-neural,warstadt2022artificial,wilcox2022using}.
To benchmark the linguistic competence of LMs, computational psycholinguists have created \textbf{targeted syntactic evaluation} benchmarks, which feature minimally-different pairs of sentences differing in grammaticality; success is measured by whether LMs assign higher probability to the grammatical sentence in each pair \citep{marvin-linzen-2018-targeted}. Despite the increasing use of LMs as models of human linguistic competence and how much easier it is to experiment on them than human brains, we do not understand the mechanisms underlying model behaviour---LMs remain largely uninterpretable.

\begin{figure}[tp]
    \centering
    \includegraphics[width=0.9\columnwidth]{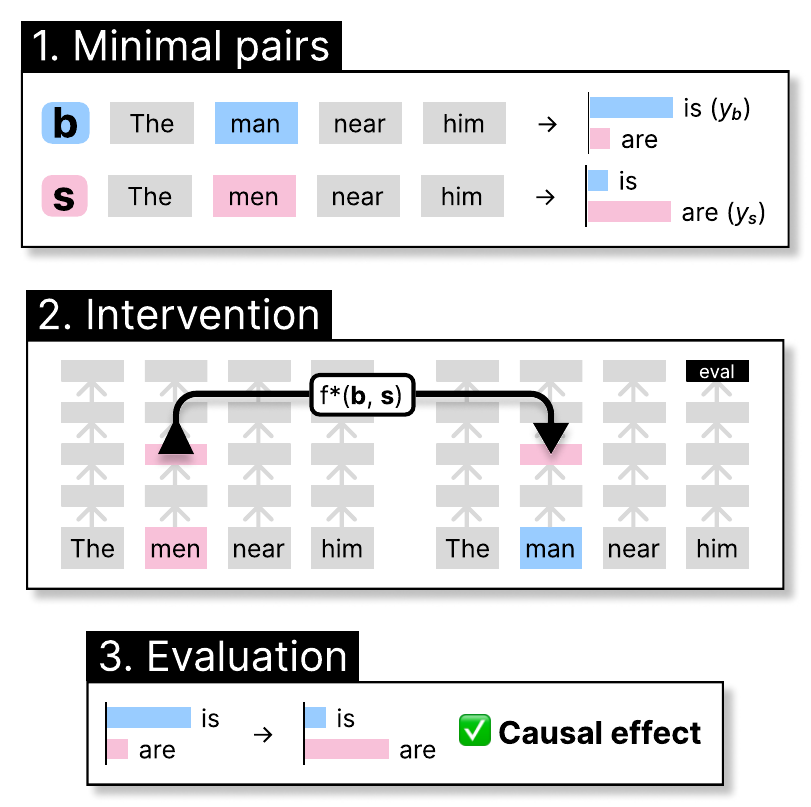}
    \caption{The \benchmarktitle{} pipeline: \textbf{(1)} take an input minimal pair ($\mathbf{b}, \mathbf{s}$) exhibiting a linguistic alternation that affects next-token predictions ($y_b, y_s$); \textbf{(2)} intervene on the base forward pass using a pre-defined intervention function that operates on aligned representations from both inputs; \textbf{(3)} check how this intervention affected the next-token prediction probabilities. In aggregate, such interventions assess the causal role of the intervened representation on the model's behaviour.}
    \label{fig:main}
\end{figure}

The \textbf{linear representation hypothesis} claims that `concepts' form linear subspaces in the representations of neural models. An increasing body of experimental evidence from models trained on language and other tasks supports this idea \citep{mikolov-etal-2013-linguistic,elhage2022superposition,park2023linear,nanda-etal-2023-emergent}. Per this hypothesis, information about high-level linguistic alternations should be localised to linear subspaces of LM activations. Methods for finding such features, and even modifying activations in feature subspaces to causally influence model behaviour, have proliferated, including probing \citep{ettinger-etal-2016-probing,adi2016fine}, distributed alignment search \citep[DAS;][]{geiger2023finding}, and difference-in-means \citep{marks2023geometry}.

Psycholinguistics and interpretability have complementary needs: thus far, psycholinguists have evaluated LMs on extensive benchmarks but neglected understanding their internal mechanisms, while new interpretability methods have only been evaluated on one-off datasets and so need better benchmarking. Thus, we introduce \benchmarktitle{} (\cref{fig:main}). We adapt linguistic tasks from SyntaxGym \citep{gauthier-etal-2020-syntaxgym} to benchmark interpretability methods on their ability to find linear features in LMs that, when subject to intervention, causally influence linguistic behaviours. We study the \texttt{pythia} family of models \citep{biderman2023pythia}, finding that DAS is the most efficacious method. However, our investigation corroborates prior findings that DAS is powerful enough to make the model produce arbitrary input--output mappings \citep{wu2023interpretability}. To address this, we adapt the notion of control tasks from the probing literature \citep{hewitt-liang-2019-designing}, finding that adjusting for performance on the arbitrary mapping task reduces the gap between DAS and other methods.

We further investigate how LMs learn two difficult linguistic behaviours during training: filler--gap extraction and negative polarity item licensing. We find that the causal mechanisms require multi-step movement of information, and that they emerge in discrete stages (not gradually) early in training.

\section{Related work}
\begin{figure*}[!t]
    \centering
    \includegraphics[width=\textwidth]{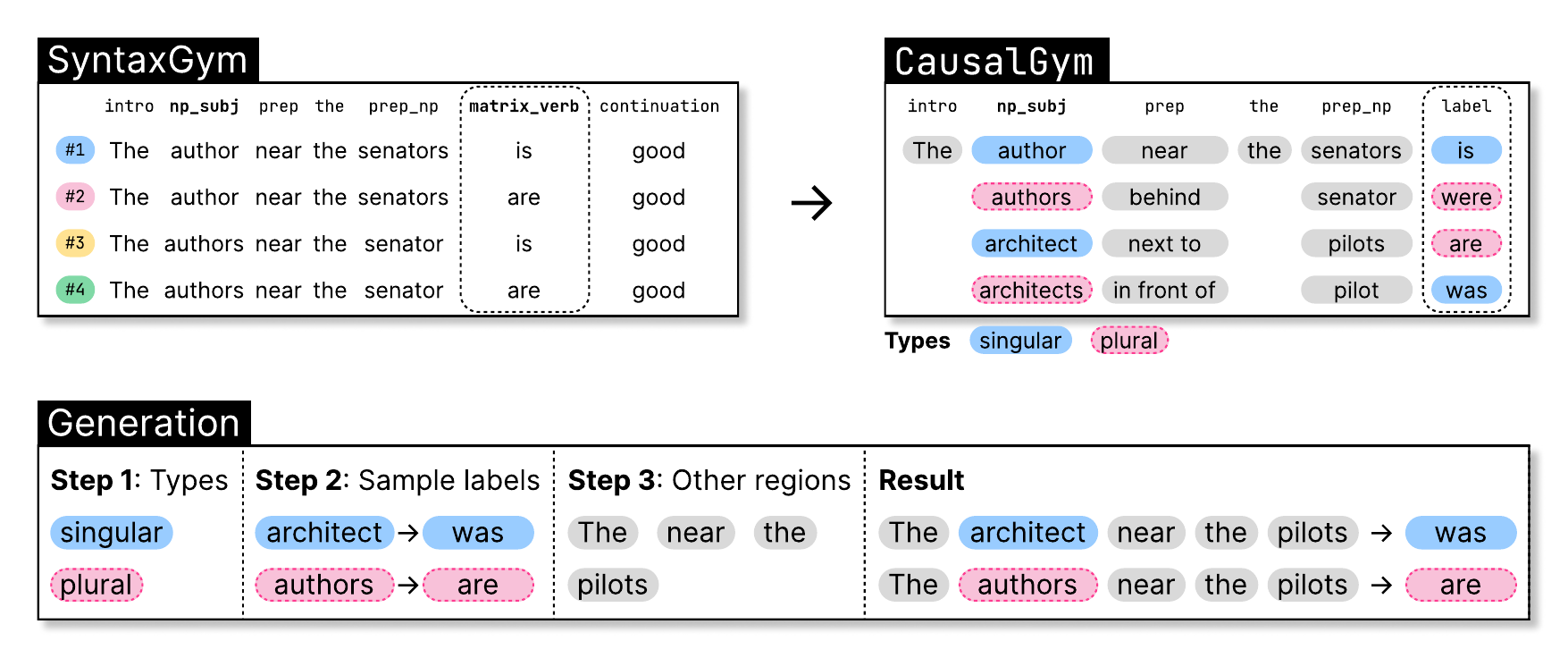}
    \caption{An example of the \benchmarktitle{} conversion process on the test suite \textit{Subject-Verb Number Agreement (with prepositional phrase)}. The left side shows how items are structured in SyntaxGym originally, which we process into the templatic format on the right. The bottom shows how we sample a minimal pair.}
    \label{fig:conversion}
\end{figure*}

\paragraph{Targeted syntactic evaluation.} Benchmarks adhering to this paradigm include SyntaxGym \citep{gauthier-etal-2020-syntaxgym,hu-etal-2020-systematic}, BLiMP \citep{warstadt-etal-2020-blimp-benchmark}, and several earlier works \citep{linzen-etal-2016-assessing,gulordava-etal-2018-colorless,marvin-linzen-2018-targeted,futrell-etal-2019-neural}.  We use the SyntaxGym evaluation sets over BLiMP even though the latter has many more examples, because we require minimal pairs that are grammatical sentences alternating along a specific feature (e.g.~number). Such pairs be templatically constructed with minimal adaptation using SyntaxGym's format.

\paragraph{Interventional interpretability.} \label{sec:intervention} Interventions are the workhorse of causal inference \citep{causality}, and have thus been adopted by recent work in interpretability for establishing the causal role of neural network components in implementing certain behaviours \citep{genderbias,causalabstraction,pmlr-v162-geiger22a,geiger2023causal,meng2022,chan2022causal,goldowsky2023localizing}, particularly linguistic ones like coreference and gender bias \citep{lasri-etal-2022-probing,wang2022interpretability,hanna-etal-2023-language,chintam-etal-2023-identifying,yamakoshi-etal-2023-causal,hao-linzen-2023-verb,chen2023sudden,amini-etal-2023-naturalistic,guerner2023geometric}.
The approach loosely falls under the nascent field of \textit{mechanistic interpretability}, which seeks to find interpretable mechanisms inside neural networks \citep{olah2022mechanistic}.

We illustrate the interventional paradigm in \cref{fig:main}; given a base input $\mathbf{b}$ and source input $\mathbf{s}$, all interventional approaches take a model-internal component $f$ and replace its output with that of $f^*(\mathbf{b}, \mathbf{s})$, which modifies the representation of $\mathbf{b}$ using that of $\mathbf{s}$. The core idea of intervention is adopted directly from the do-operator used in causal inference; we test the intervention's effect on model output to establish a causal relationship.

\section{Benchmark}

To create \benchmarktitle{}, we converted the core test suites in SyntaxGym \citep{gauthier-etal-2020-syntaxgym} into templates for generating large numbers of span-aligned minimal pairs, a process we describe below along with our evaluation setup.

\subsection{Premise}

Each test suite in SyntaxGym focuses on a single linguistic feature, constructing English-language minimal pairs that minimally adjust that feature to change expectations about how a sentence should continue. A test suite contains several \textit{items} which share identical settings for irrelevant features, and each item has some \textit{conditions} which vary only the important feature. All items adhere to the same templatic structure, sharing the same ordering and set of \textit{regions} (syntactic units). To measure whether models match human expectations, SyntaxGym evaluates the model's surprisal at specific regions between differing conditions.

For example, the \textit{Subject-Verb Number Agreement (with prepositional phrase)} task constructs items consisting of 4 conditions, which set all possible combinations of the number feature on subjects and their associated verbs, as well as the \textit{opposite} feature on a distractor noun. Each example in this test suite follows the template
\begin{examples}
    \item The \token{\texttt{np\_subj}} \token{\texttt{prep}} the \token{\texttt{prep\_np}} \token{\texttt{matrix\_verb}} \token{\texttt{continuation}}.
\end{examples}
where, in a single item, the regions \texttt{np\_subj} and \texttt{matrix\_verb} are modified along the number feature, and \texttt{prep\_np} is a distractor. For example:
\begin{examples}\setlength{\itemsep}{0pt}
    \item{The \token{\textbf{author}} \token{near} the \token{senators} \token{\textbf{is}} \token{good}.\label{ex:match_sing}}
    \item\sqz{*}{The \token{\textbf{author}} \token{near} the \token{senators} \token{\textbf{are}} \token{good}.\label{ex:mismatch_sing}}
    \item\sqz{*}{The \token{\textbf{authors}} \token{near} the \token{senator} \token{\textbf{is}} \token{good}.\label{ex:mismatch_plural}}
    \item{The \token{\textbf{authors}} \token{near} the \token{senator} \token{\textbf{are}} \token{good}.\label{ex:match_plural}}
\end{examples}
Humans expect agreement between the number feature on the verb and the subject, as in \eg{ex:match_sing} and \eg{ex:match_plural}. 
On this test suite, SyntaxGym measures if the surprisal at the verb satisfies the following inequalities between conditions: $p(\text{is} \mid \text{author}) > p(\text{are} \mid \text{author})$ and $p(\text{are} \mid \text{authors}) > p(\text{is} \mid \text{authors})$.

\subsection{Templatising SyntaxGym}
\label{sec:template}

Our goal is to study how LMs implement mechanisms for converting feature alternations in the input into corresponding alternations in the output---e.g., how does an LM keep track of the number feature on the subject when it needs to output an agreeing verb?
In adapting SyntaxGym for this purpose, we need to address two issues: (1) to study model mechanisms, we only want \textit{grammatical} pairs of sentences; and (2) SyntaxGym test suites contain $<50$ items, while we need many more for training supervised interpretability methods and creating non-overlapping test sets.

Thus, we select the two grammatical conditions from each item and simplify the behaviour of interest into an explicit input--output mapping. For example, we recast \textit{Subject-Verb Number Agreement (with prepositional phrase)} into counterfactual pairs that elicit singular or plural verbs based on the number feature of the subject, and hold everything else (including the distractor) constant:
\begin{examples}
    \item  \label{ex:pair} 
    \begin{examples}\setlength{\itemsep}{0pt}
    \item The \textbf{author} near the senators $\Rightarrow$ \textbf{is} \label{ex:pair1}
    \item The \textbf{authors} near the senators $\Rightarrow$ \textbf{are}
    \end{examples}
\end{examples}

To be able to generate many examples for training, we use the aligned regions as slots in a template that we can mix-and-match between items to combinatorially generate pairs, illustrated in \cref{fig:conversion}. We manually removed options that would have resulted in questionably grammatical sentences.

For generation using our format, each template has a set of types $T$ which govern the input \textbf{label variable} and the expected next-token prediction \textbf{label}. To generate a counterfactual pair, we first sample two types $t_1, t_2 \sim T$ such that $t_1 \neq t_2$. Then, for the label variable and label, we sample an option of that type $t_1$ (for the first sentence) or $t_2$ (for the second). Finally, for the non-label variable regions, we sample one option and set both sentences to that. In \cref{fig:conversion}, we show the generation process in the bottom panel; types for the label variable and label options are colour-coded.

\begin{figure*}[tp]
    \centering
    \includegraphics[width=\textwidth]{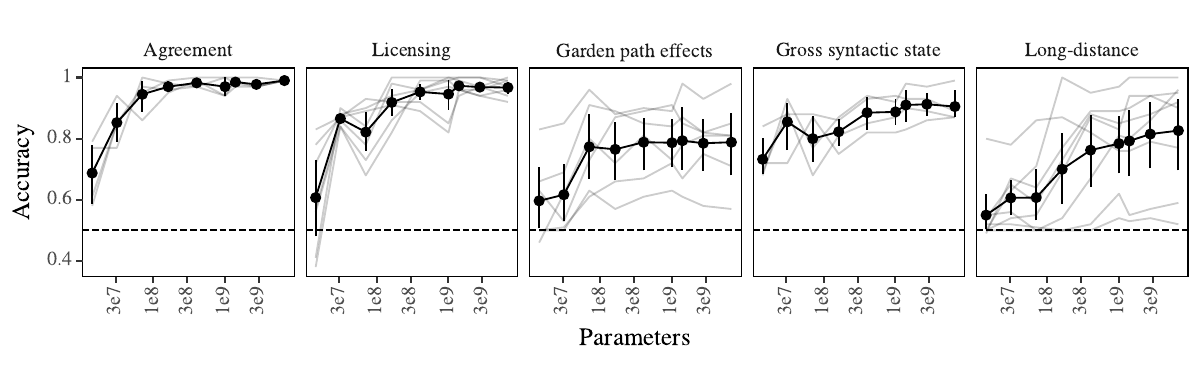}
    \caption{Accuracy of \texttt{pythia}-family models on the \benchmarktitle{} tasks, grouped by type, with scale. The dashed line is random-chance accuracy (50\%).}
    \label{fig:acc}
\end{figure*}

\subsection{Tasks}

\benchmarktitle{} contains 29 tasks, of which one is novel (\texttt{agr\_gender}) and 28 were templatised from SyntaxGym. Of the 33 test suites in the original release of SyntaxGym, we only used tasks from which we could generate paired grammatical sentences (leading us to discard the 2 center embedding tasks), and merged the 6 gendered reflexive licensing tasks into 3 non-gendered ones. We show task accuracy vs.~model scale in \cref{fig:acc}. Examples of pairs generated for each task are provided in \cref{sec:tasks}.

\subsection{Evaluation}
\label{sec:eval}

An evaluation sample consists of a base input $\mathbf{b}$, source input $\mathbf{s}$, ground-truth base label $y_b$, and ground-truth source label $y_s$. For example, the components of \eg{ex:pair} are\looseness=-1\vspace{-0.5em}
\begin{examples}
    \item $\underbrace{\text{The author near the senators}}_{\mathbf{b}} \Rightarrow \underbrace{\text{is}}_{y_b}$\vspace{-0.5em}
    \item $\underbrace{\text{The authors near the senators}}_{\mathbf{s}} \Rightarrow \underbrace{\text{are}}_{y_s}$\vspace{-0.5em}
\end{examples}\looseness=-1
A successful intervention will take the original LM $p$ running on input $\mathbf{b}$ and make it predict $y_s$ as the next token. We measure the strength of an intervention by its \textbf{log odds-ratio}.

First, we select a component $f$, which can be any part of a neural network that outputs a representation, inside the model $p$. When the model is run on input $\mathbf{b}$, this component produces a representation we denote $f(\mathbf{b})$.
We perform an intervention which replaces the output of $f$ with an output of $f^*$ as in \cref{sec:intervention}. To produce a representation, $f^*$ may modify the base representation with reference to the source representation, and so its output is $f^*(\mathbf{b}, \mathbf{s})$. The intervention results in an intervened language model which we denote informally as $\intlm$. In the framework of causal abstraction \citep{causalabstraction}, if this intervention successfully makes the model behave as if its input was $\mathbf{s}$, then the representation at $f$ is causally aligned with the high-level linguistic feature alternating in $\mathbf{b}$ and $\mathbf{s}$.

We now operationalise a measure of causal effect.
Taking the original model $p$, the intervened model $\intlm$, and the evaluation sample, we define the log odds-ratio as:
\begin{align}
    \mathsf{Odds}&(p, \intlm, \langle \mathbf{b}, \mathbf{s}, y_b, y_s \rangle)\nonumber\\
    &= \log{\left(\frac{p(y_b \mid \mathbf{b})}{p(y_s \mid \mathbf{b})} \cdot \frac{\intlm(y_s \mid \mathbf{b}, \mathbf{s})}{\intlm(y_b \mid \mathbf{b}, \mathbf{s})}\right)}
\end{align}
where a greater log odds-ratio indicates a larger causal effect at that intervention site, and a log odds-ratio of $0$ indicates no causal effect. Given an evaluation set $E$, the average log odds-ratio is
\begin{align}
    \mathsf{AvgOdds}&(p, \intlm, E) = \nonumber\\
    &\frac{1}{\lvert E \rvert} \sum_{e \in E} \mathsf{Odds}(p, \intlm, e) \label{eq:avg_odds}
\end{align}

\section{Methods}

We briefly describe our choice of $f^*$ and the feature-finding methods that we benchmark in this paper.

\subsection{Preliminaries}
In this paper, we only benchmark interventions along a single feature direction, i.e.~one-dimensional distributed interchange intervention \citep[1D DII;][]{geiger2023finding}.
DII is an interchange intervention that operates on a non-basis-aligned subspace of the activation space.
Formally, given a feature vector $\mathbf{a} \in \mathbb{R}^n$ and $f$, 1D DII defines $f^*$ as
\begin{equation}
    f^*_{\mathbf{a}}(\mathbf{b}, \mathbf{s}) = f(\mathbf{b}) + (f(\mathbf{s})\mathbf{a}^\top - f(\mathbf{b})\mathbf{a}^\top)\mathbf{a}\label{eq:dii}
\end{equation}
As noted above, when our intervention replaces $f$ with $f^*_{\mathbf{a}}$, we denote the new model as $\diilm$.

We fix $f$ to operate on token-level representations; since $\mathbf{b}$ and $\mathbf{s}$ may have different lengths due to tokenisation; we align representations at the last token of each template region.

In principle, we allow future work to consider other forms of $f^*$, but 1D DII has two useful properties. Given the linear representation hypothesis and that \benchmarktitle{} exclusively studies binary linguistic features, 1D DII ought to be sufficiently expressive for controlling model behaviour. Furthermore, probes trained on binary classification tasks operate on a one-dimensional subspace of the representation, and thus we can directly use the weight vector of a probe as the parameter $\mathbf{a}$ in \cref{eq:dii}---\citet{tigges2023linear} used a similar setup to causally evaluate probes.

We study seven methods, of which four are supervised: distributed alignment search (DAS), linear probing, difference-in-means, and LDA. The other three are unsupervised: PCA, $k$-means, and (as a baseline) sampling a random vector. All of these methods provide us a feature direction $\mathbf{a}$ that we use as a constant in \cref{eq:dii}. For probing and unsupervised methods, we use implementations from \texttt{scikit-learn} \citep{sklearn}. To train distributed alignment search and run 1D DII, we use the \texttt{pyvene} library \citep{wu2024pyvene}. Further training details are in \cref{sec:training_details}. We formally describe each method below.

\subsection{Definitions}

\begin{table*}[tp]
    \small
    \centering
    \adjustbox{width=\textwidth}{
    \begin{tabular}{rrrrrrrrrrcrrrrrr}
        \toprule
        \multirow{2}{*}{\textbf{Mod.}} & \multirow{2}{*}{\textbf{Acc.}} & \multicolumn{7}{c}{\textbf{Overall odds-ratio} ($\uparrow$)} & & \multicolumn{7}{c}{\textbf{Selectivity} ($\uparrow$)} \\
        \cmidrule{3-9} \cmidrule{11-17}
        & & \textbf{DAS} & \textbf{Probe} & \textbf{Mean} & \textbf{PCA} & \textbf{$k$-m.} & \textbf{LDA} & \textbf{Rand.} & & \textbf{DAS} & \textbf{Probe} & \textbf{Mean} & \textbf{PCA} & \textbf{$k$-m.} & \textbf{LDA} & \textbf{Rand.} \\
        \midrule
        \texttt{14m} & 0.62 & \textbf{3.94} & 1.16 & 1.04 & 0.48 & 0.50 & 0.11 & 0.03 & & \textbf{1.84} & 1.38 & 1.24 & 0.54 & 0.55 & 0.15 & 0.08 \\
        \texttt{31m} & 0.74 & \textbf{5.82} & 2.22 & 1.80 & 0.83 & 0.85 & 0.08 & 0.02 & & \textbf{2.75} & 2.63 & 2.03 & 0.86 & 0.88 & 0.13 & 0.03 \\
        \texttt{70m} & 0.77 & \textbf{7.60} & 2.70 & 2.12 & 1.16 & 1.20 & 0.11 & 0.03 & & \textbf{2.87} & 2.86 & 2.15 & 1.05 & 1.09 & 0.16 & 0.05 \\
        \texttt{160m} & 0.82 & \textbf{7.93} & 3.13 & 2.23 & 1.26 & 1.29 & 0.12 & 0.02 & & 2.93 & \textbf{3.27} & 2.34 & 1.24 & 1.26 & 0.15 & 0.04 \\
        \texttt{410m} & 0.86 & \textbf{10.24} & 3.69 & 3.22 & 2.15 & 2.19 & 0.34 & 0.05 & & 3.96 & \textbf{4.20} & 3.33 & 2.07 & 2.12 & 0.43 & 0.06 \\
        \texttt{1b} &  0.86 & \textbf{10.74} & 3.66 & 3.17 & 2.07 & 2.13 & 0.29 & 0.03 & & 3.34 & \textbf{4.24} & 3.09 & 1.78 & 1.85 & 0.36 & 0.04 \\
        \texttt{1.4b} & 0.88 & \textbf{9.58} & 3.48 & 3.06 & 1.96 & 2.02 & 0.37 & 0.02 & & 2.99 & \textbf{4.08} & 3.21 & 1.87 & 1.94 & 0.46 & 0.03 \\
        \texttt{2.8b} & 0.88 & \textbf{8.88} & 3.72 & 3.19 & 1.93 & 2.00 & 0.31 & 0.01 & & 2.57 & \textbf{4.15} & 3.31 & 1.69 & 1.75 & 0.39 & 0.01 \\
        \texttt{6.9b} & 0.89 & \textbf{9.95} & 3.42 & 2.91 & 1.81 & 1.87 & 0.27 & 0.01 & &
        2.48 & \textbf{3.79} & 2.85 & 1.50 & 1.54 & 0.34 & 0.02 \\
        \bottomrule
    \end{tabular}}
    \caption{Overall odds-ratio (\cref{sec:efficacious}) and selectivity (\cref{sec:control}) of each feature-finding method averaged over all tasks in \benchmarktitle{}. We also report average task accuracy, which increases with scale. For models larger than \texttt{pythia-70m}, we report the better of two probes trained with different hyperparameters (\cref{sec:hyperparam}).}
    \label{tab:odds}
\end{table*}

\paragraph{DAS.}
Given a training set $T$, we learn the intervention direction, potentially distributed across many neurons, that maximises the output probability of the counterfactual label.
Formally, we first randomly initialise $\feature{das}$ and intervene on the model $p$ with it to get $\diilmf{das}$. We freeze the model weights and optimise $\feature{das}$ such that we minimise the cross-entropy loss with the target output $y_s$:
\begin{equation}
\min_{\feature{das}}\left\{\;- \smashoperator{\sum_{\langle \mathbf{b}, \mathbf{s}, y_b, y_s \rangle \in T}} 
\log{\diilmf{das}(y_s \mid \mathbf{b}, \mathbf{s})}\right\}
\end{equation}
The learned DAS parameters $\feature{das}$ then define a function $f^*_{\feature{das}}$ using \eqref{eq:dii}.

\paragraph{Linear probe.} Linear probing classifiers have been the dominant feature-finding method for neural representations of language \citep{belinkov-2022-probing}. A probe outputs a distribution over classes given a representation $\mathbf{x} \in \mathbb{R}^n$:
\begin{equation}
q_{\boldsymbol{\theta}}(y \mid \mathbf{x}) = \text{softmax}(\feature{probe}\cdot f(\mathbf{x}) + b)
\end{equation}
We learn the parameters $\boldsymbol{\theta}$ of the probe over the base training set examples (so, maximising $q_{\boldsymbol{\theta}}(y_b \mid \mathbf{b})$) using the SAGA solver \citep{defazio2014saga} as implemented in \texttt{scikit-learn}, and the parameters $\feature{probe}$ define the intervention function $f^*_{\feature{probe}}$.

\paragraph{Diff-in-means.} The difference in per-class mean activations has been surprisingly effective for controlling representations \citep{marks2023geometry,li2023inference} and erasing linear features \citep{belrose2023leace,belrose2023diff}. To implement this approach, we take the base input--output pairs $\langle \mathbf{b}, y_b \rangle$ from the training set $T$, where $y_b \in \{y_1, y_2\}$, and group them by the identity of their labels. Thus, we have $X_1 = \{\mathbf{b} \in T : y_b = y_1 \}$ and $X_2 = \{\mathbf{b} \in T : y_b = y_2 \}$. The diff-in-means method is then defined as follows:
\begin{align}
\feature{mean} &= \frac{1}{\lvert X_1 \rvert}\sum_{\mathbf{x} \in X_1}f(\mathbf{x}) - \frac{1}{\lvert X_2 \rvert}\sum_{\mathbf{x} \in X_2}f(\mathbf{x})\\
&= \boldsymbol{\mu}_1 - \boldsymbol{\mu}_2
\end{align}
and as usual $\feature{mean}$ defines the function $f^*_{\feature{mean}}$.

\paragraph{Linear disciminant analysis.} LDA assumes that each class is distributed according to a Gaussian and all classes share the same covariance matrix $\mathbf{\Sigma}$. Given the per-class means $\boldsymbol{\mu}_1$ and $\boldsymbol{\mu}_2$,
\begin{equation}
\feature{lda} = \mathbf{\Sigma}^{-1}(\boldsymbol{\mu}_1 - \boldsymbol{\mu}_2)
\end{equation}

\paragraph{Principal component analysis (PCA).} We intervene along the first principal component, which is a vector $\feature{pca}$ that maximises the variance in mean-centered activations (denoted $\widetilde{f}(\mathbf{x})$).
\begin{equation}
\max_{\feature{pca}}\left\{\sum_{\mathbf{x} \in X_1 \cup X_2}\left(\widetilde{f}(\mathbf{x}) \cdot \feature{pca}\right)^2\right\}
\end{equation}
PCA was previously used to debias gendered word embeddings by \citet{bolukbasi2016man}.

\paragraph{$k$-means.} We use 2-means and learn a clustering of activations into two sets $S_1, S_2$ that minimises the variance of the activations relative to their class centroids $\boldsymbol{\mu}_1, \boldsymbol{\mu}_2$. Our feature direction is
\begin{equation}
\feature{kmeans} = \boldsymbol{\mu}_1 - \boldsymbol{\mu}_2
\end{equation}

\section{Experiments}
\label{sec:experiments}

We perform all experiments on the \texttt{pythia} model series \citep{biderman2023pythia}, which includes 10 models ranging from 14 million to 12 billion parameters, all trained on the same data in the same order. This model series allows us to study how feature representations change with scale and training data size in a controlled manner---all models were trained on the same data in the same order, and checkpoints are provided.

\subsection{Measuring causal efficacy}
\label{sec:efficacious}

The Transformer \citep{Vaswani-etal:2017} is organised around the \textbf{residual stream} \citep{elhage2021mathematical}, which each attention and MLP layer reads from and additively writes to. The residual stream is an information bottleneck; information from the input must be present at some token in every layer's residual stream in order to reach the next layer and ultimately affect the output.

Therefore, given a feature  present in the input and influencing the model output, we should be able to find a causally-efficacious subspace encoding that feature in at least one token position in every layer. If the feature is binary (such as the ones we study in \benchmarktitle{}), then 1D DII should be sufficient for this.

Thus, for each task in \benchmarktitle{}, we take the function of interest $f$ to be the state of the residual stream after the operation of a Transformer layer $l \in L$ at the last token of a particular region $r \in R$. For notational convenience, we denote this function as $f^{(l,r)}$. We learn 1D DII using each method $m$ for every such function. We use a trainset $T$ of 400 examples for each benchmark task, and evaluate on a non-overlapping set $E$ of 100 examples.\footnote{Further training details are given in \cref{sec:training_details}, and we report hyperparameter tuning experiments on a dev set in \cref{sec:hyperparam}.} Each such experiment results in an intervened model that we denote $p_{f^{(l,r)} \gets f^*_{\mathbf{a}_m}}$. To compute the overall log odds-ratio for a feature-finding method on a particular model on a single task, we take the maximum of the average odds-ratio (\cref{sec:eval}) over regions at a specific layer, and then average over all layers:
\begin{equation}
    \mathsf{OverallOdds}(p, m, E) \label{eq:overall_odds}\\
\end{equation}
\begin{equation}
    = \frac{1}{\lvert L \rvert} \sum_{l \in L}  \left( \max_{r \in R} \left(\mathsf{AvgOdds}(p, p_{f^{(l,r)} \gets f^*_{\mathbf{a}_m}}, E) \right) \right) \nonumber
\end{equation}
This metric rewards a method for finding a highly causally-efficacious region in every layer.

\begin{figure*}
    \centering
    \includegraphics[width=\textwidth]{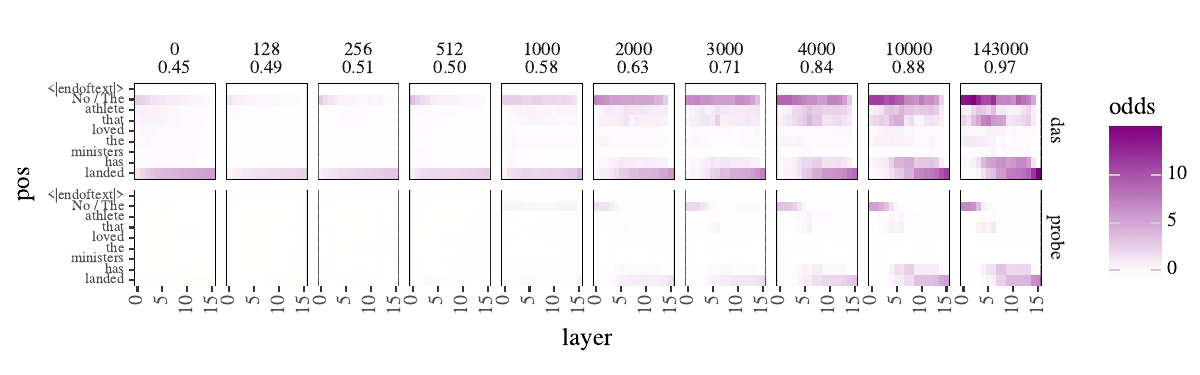}
    \caption{Odds-ratio for checkpoints of \texttt{pythia-1b} on the task \texttt{npi\_any\_subj\_relc}, plotted at every layer and template region. The $y$-axis is labelled with an example pair of sentences. The plot titles are labelled with the checkpoint and task accuracy. Darker regions indicate a token in a specific layer where causal effect was high.}
    \label{fig:npi}
\end{figure*}

\subsection{Controlling for expressivity}
\label{sec:control}

DAS is the the only method with a causal training objective. Other methods do not optimise for, or even have access to, downstream model behaviour. 
\citet{wu2023interpretability} found that a variant of DAS achieves substantial causal effect even on a randomly-initialised model or with irrelavant next-token labels, both settings where no causal mechanism should exist. How much of the causal effect found by DAS is due to its expressivity? Research on probing has faced a similar concern: to what extent is a probe's accuracy due to its expressivity rather than any aspect of the representation being studied? \citet{hewitt-liang-2019-designing} propose comparing to accuracy on a \textbf{control task} that requires memorising an input-to-label mapping.

We adapt this notion to \benchmarktitle{}, introducing control tasks where the next-token labels $y_b, y_s$ are mapped to the arbitrary tokens `\_dog' and `\_give' while preserving the class partitioning.\footnote{The input-to-label mapping in \benchmarktitle{} tasks is dependent on the input token types, so we cannot exactly replicate \citeauthor{hewitt-liang-2019-designing}.
The setup we instead use is from \citet{wu2023interpretability}.} For example, on the gender-agreement task \texttt{agr\_gender}, we replace the label `\_he' with `\_dog' and `\_she' with `\_give'. We define selectivity for each method by taking the difference between odds-ratios on the original task and the control task for each $f$, and then compute the overall odds-ratio as in \cref{eq:overall_odds}.

\subsection{Results}

\begin{figure}[t]
\centering
\begin{subfigure}[b]{0.47\linewidth}
\centering
\includegraphics[height=1.6in]{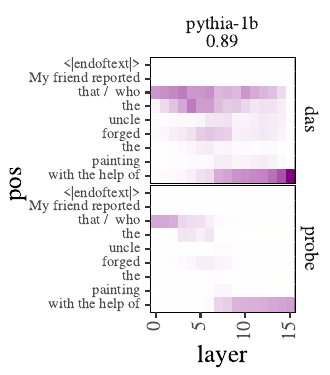}
\caption{\texttt{filler\_gap\_subj}}
\label{fig:fillergap_final}
\end{subfigure}
\hfill 
\begin{subfigure}[b]{0.47\linewidth}
\centering
\includegraphics[height=1.6in]{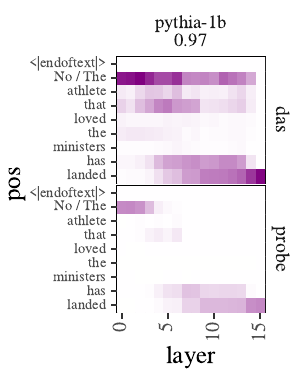}
\caption{\texttt{npi\_any\_subj\_relc}}
\label{fig:npi_final}
\end{subfigure}
\caption{Odds-ratio for each layer and region using DAS and probing on \texttt{pythia-1b}, on two tasks.}
\label{fig:casestudies}
\end{figure}

We summarise the results for each method in \cref{tab:odds} by reporting overall odds-ratio and selectivity averaged over all tasks for each model. For a breakdown, see \cref{sec:overall_odds_task,sec:overall_odds_layer}.

We find that DAS consistently finds the most causally-efficacious features.
The second-best method is probing, followed by difference-in-means. The unsupervised methods PCA and $k$-means are considerably worse. Despite supervision, LDA barely outperforms random features.

However, DAS is not considerably more selective or (at larger scales) even less selective than probing or diff-in-means; it can perform well on arbitrary input--output mappings. This suggests that its access to the model outputs during training is responsible for much of its advantage.

\section{Case studies}
\label{sec:studies}

In this section, we use \benchmarktitle{} to study how LMs learn negative polarity item (NPI) licensing and wh-extraction from prepositional phrases over the course of training using checkpoints of \texttt{pythia-1b}. We first describe the tasks.

\paragraph{\texttt{npi\_any\_subj-relc}.} NPIs are lexemes that can only occur in negative-polarity sentential contexts. In this task, we specifically check whether the NPI \textit{any} is correctly licensed by a negated subject, giving minimal pairs like~\looseness=-1\vspace{-0.1in}
\begin{examples}\setlength{\itemsep}{0pt}
    \item \textbf{No} athlete that loved the ministers has landed $\Rightarrow$ \textbf{any}
    \item \textbf{The} athlete that loved the ministers has landed $\Rightarrow$ \textbf{some}\label{ex:ppi}
\end{examples}\vspace{-0.1in}
In \eg{ex:ppi}, where there is no negation at the sentence level, it would be ungrammatical to continue the sentence with the NPI \textit{any}.

\paragraph{\texttt{filler\_gap\_subj}.} Filler--gap dependencies in English occur when interrogatives are extracted out of and placed in front of a clause. The position from which they are extracted must remain empty. The task \texttt{filler\_gap\_subj} requires an LM to apply this rule when extracting from a distant prepositional phrase, e.g.~\vspace{-0.1in}
\begin{examples}\setlength{\itemsep}{0pt}
    \item My friend reported \textbf{that} the uncle forged the painting with the help of $\Rightarrow$ \textbf{him}
    \item My friend reported \textbf{who} the uncle forged the painting with the help of $\Rightarrow$ \textbf{.}\label{ex:gap}\vspace{-0.1in}
\end{examples}
In \eg{ex:gap}, it would be ungrammatical for the preposition to have an explicit object since \textit{who} was extracted from that position, leaving behind a gap.

\paragraph{Final mechanisms.}

\begin{figure*}
    \centering
    \includegraphics[width=\textwidth]{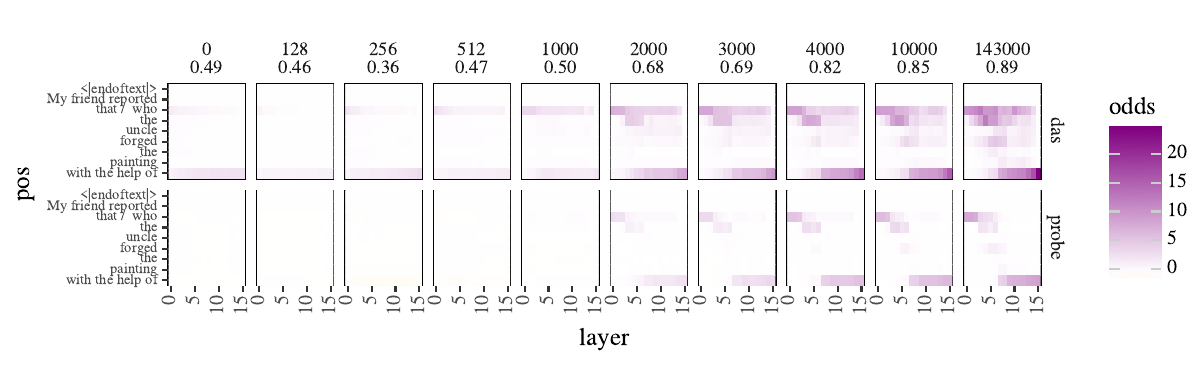}
    \caption{Odds-ratio for checkpoints of \texttt{pythia-1b} on the task \texttt{filler\_gap\_subj}, plotted at every layer and template region.}
    \label{fig:filler}
\end{figure*}

We use the experimental setup of \cref{sec:efficacious} and plot the average odds-ratio for each region and layer on the final checkpoint of \texttt{pythia-1b} in \cref{fig:casestudies}. For both tasks, we find that the input feature crosses over several different positions before arriving at the output position. For example, in the NPI mechanism (\cref{fig:npi_final}), the negation feature is moved to the complementiser \textit{that} in the early layer, into the auxiliary verb at middle layers, and the main verb in later layers, where its presence is used to predict the NPI \textit{any}. The filler--gap mechanism is similarly complex.

\subsection{Training dynamics}
\label{sec:npi}

To study how the mechanisms emerge over the course of training, we run the exact same experiments on earlier checkpoints of \texttt{pythia-1b}.

\paragraph{\texttt{npi\_any\_subj-relc}.} In \cref{fig:npi}, the effect first emerges at the NPI (all but last layer) and the main verb (step 1000), then abruptly the auxiliary becomes important at middle layers and the NPI effect is pushed down to early layers (step 2000), and finally another intermediate locations is added at \textit{that} (step 3000). The effect is also distributed across multiple regions in the intermediate layers.

\paragraph{\texttt{filler\_gap\_subj}.} This behaviour takes longer to learn than NPI licensing (\cref{fig:filler}). The mechanism emerges in two stages: at step 2000, it includes the filler position (\textit{that} / \textit{who}), the first determiner \textit{the}, and the final token. After step 10K, the main verb is  added to the mechanism.

\paragraph{Discussion.} For both tasks, the model intially learns to move information directly from the alternating token to the output position. Later in training, intermediate steps are added in the middle layers. 
DAS finds a greater causal effect across the board, but both methods largely agree on which regions are the most causally efficacious at each layer. Notably, DAS finds causal effect at all timesteps, even when the model has just been initialised; this corroborates \citeposs{wu2023interpretability} findings.

\section{Conclusion}

We introduced \benchmarktitle{}, a multi-task benchmark of linguistic behaviours for measuring the causal efficacy of interpretability methods. We showed the impressive performance of distributed alignment search, but also adapted a notion of control tasks to causal evaluation to enable fairer comparison of methods. Finally, we studied how causal effect propagates in training on two linguistic tasks: NPI licensing and filler--gap dependency tracking.

In recent years, much effort has been devoted towards developing causally-grounded methods for understanding neural networks. A probe achieving high classification accuracy provides no guarantee that the model actually distinguishes those classes in downstream computations; evaluating probe directions for causal effect is an intuitive test for whether they reflect features that the model uses downstream. Overall, while methods may come and go, we believe the causal evaluation paradigm will continue to be useful for the field.

A major motivation for releasing \benchmarktitle{} is to encourage computational psycholinguists to move beyond studying the input--output behaviours of LMs. Our case studies in \cref{sec:studies} are a basic example of the analysis that new methods permit. Ultimately, understanding how LMs learn linguistic behaviours may offer insights into fundamental properties of language (cf.~\citealp{kallini2024mission,wilcox2022using}).

We hope that \benchmarktitle{} will encourage comprehensive evaluation of new interpretability methods and spur adoption of the interventional paradigm in computational psycholinguistics.

\section*{Limitations}

While \benchmarktitle{} includes a range of linguistic tasks, there are many non-linguistic behaviours on which we may want to use interpretability methods, and so we encourage future research on a greater variety of tasks. In addition, \benchmarktitle{} includes only English data, and comparable experiments with other languages might yield substantially different results, thereby providing us with a much fuller picture of the causal mechanisms that LMs learn to use. Furthermore, results may differ on other models, since models in the \texttt{pythia} series were trained on the same data in a fixed order; different training data may result in different mechanisms. Finally, justified by the nature of our tasks, we only benchmark methods that operate on one-dimensional linear subspaces; multi-dimensional linear methods as well as non-linear ones await being benchmarked.

\section*{Ethics statement}

Interpretability is a rapidly-advancing field, and our benchmark results render us optimistic about our ability to someday understand the mechanisms inside complex neural networks. However, successful interpretability methods could be used to justify deployment of language models in high-risk settings (e.g.~to autonomously make decisions about human beings) or even manipulate models to produce harmful outputs. Understanding a model does not mean that it is safe to use in every situation, and we caution model deployers and users against uncritical trust in models even if they are found to be interpretable.

\section*{Acknowledgements}
We thank Atticus Geiger, Jing Huang, Harshit Joshi, Jordan Juravsky, Julie Kallini, Chenglei Si, Tristan Thrush, and Zhengxuan Wu for helpful discussion about the project and their comments on the manuscript.

\bibliography{short}
\onecolumn
\appendix

\section{Tasks}
\label{sec:tasks}
\begin{table}[!ht]
    \small
    \centering
    \adjustbox{width=\textwidth}{
    \begin{tabular}{lp{0.7\textwidth}}
        \toprule
        \textbf{Task} & \textbf{Example} \\
        \midrule
        \textbf{\textit{Agreement}} (4) \\
        \texttt{agr\_gender} & \pair{John}{Jane} walked because \pair{he}{she} \\
        \texttt{agr\_sv\_num\_subj-relc} & The \pair{guard}{guards} that hated the manager \pair{is}{are} \\
        \texttt{agr\_sv\_num\_obj-relc} & The \pair{guard}{guards} that the customers hated \pair{is}{are} \\
        \texttt{agr\_sv\_num\_pp} & The \pair{guard}{guards} behind the managers \pair{is}{are} \\
        \midrule
        \textbf{\textit{Licensing}} (7) \\
        \texttt{agr\_refl\_num\_subj-relc} & The \pair{farmer}{farmers} that loved the actors embarrassed \pair{himself}{themselves} \\
        \texttt{agr\_refl\_num\_obj-relc} & The \pair{farmer}{farmers} that the actors loved embarrassed \pair{himself}{themselves} \\
        \texttt{agr\_refl\_num\_pp} & The \pair{farmer}{farmers} behind the actors embarrassed \pair{himself}{themselves} \\
        \texttt{npi\_any\_subj-relc} & \pair{No}{The} consultant that has helped the taxi driver has shown \pair{any}{some} \\
        \texttt{npi\_any\_obj-relc} & \pair{No}{The} consultant that the taxi driver has helped has shown \pair{any}{some} \\
        \texttt{npi\_ever\_subj-relc} & \pair{No}{The} consultant that has helped the taxi driver has \pair{ever}{never} \\
        \texttt{npi\_ever\_obj-relc} & \pair{No}{The} consultant that the taxi driver has helped has \pair{ever}{never} \\
        \midrule
        \textbf{\textit{Garden path effects}} (6) \\
        \texttt{garden\_mvrr} & The infant \pair{who was}{$\varnothing$} brought the sandwich from the kitchen \pair{by}{.} \\
        \texttt{garden\_mvrr\_mod} & The infant \pair{who was}{$\varnothing$} brought the sandwich from the kitchen with a new microwave \pair{by}{.} \\
        \texttt{garden\_npz\_obj} & While the students dressed \pair{,}{$\varnothing$} the comedian \pair{was}{for} \\
        \texttt{garden\_npz\_obj\_mod} & While the students dressed \pair{,}{$\varnothing$} the comedian who told bad jokes \pair{was}{for} \\
        \texttt{garden\_npz\_v-trans} & As the criminal \pair{slept}{shot} the woman \pair{was}{for} \\
        \texttt{garden\_npz\_v-trans\_mod} & As the criminal \pair{slept}{shot} the woman who told bad jokes \pair{was}{for} \\
        \midrule
        \textbf{\textit{Gross syntactic state}} (4) \\
        \texttt{gss\_subord} & \pair{While the}{The} lawyers lost the plans \pair{they}{.} \\
        \texttt{gss\_subord\_subj-relc} & \pair{While the}{The} lawyers who wore white lab jackets studied the book that described several advances in cancer therapy \pair{,}{.} \\
        \texttt{gss\_subord\_obj-relc} & \pair{While the}{The} lawyers who the spy had contacted repeatedly studied the book that colleagues had written on cancer therapy \pair{,}{.} \\
        \texttt{gss\_subord\_pp} & \pair{While the}{The} lawyers in a long white lab jacket studied the book about several recent advances in cancer therapy \pair{,}{.} \\
        \midrule
        \textbf{\textit{Long-distance dependencies}} (8) \\
        \texttt{cleft} & What the young man \pair{did}{ate} was \pair{make}{for}\\
        \texttt{cleft\_mod} & What the young man \pair{did}{ate} after the ingredients had been bought from the store was \pair{make}{for} \\
        \texttt{filler\_gap\_embed\_3} & I know \pair{that}{what} the mother said the friend remarked the park attendant reported your friend sent \pair{him}{.} \\
        \texttt{filler\_gap\_embed\_4} & I know \pair{that}{what} the mother said the friend remarked the park attendant reported the cop thinks your friend sent \pair{him}{.} \\
        \texttt{filler\_gap\_hierarchy} & The fact that the brother said \pair{that}{who} the friend trusted \pair{the}{was} \\
        \texttt{filler\_gap\_obj} & I know \pair{that}{what} the uncle grabbed \pair{him}{.} \\
        \texttt{filler\_gap\_pp} & I know \pair{that}{what} the uncle grabbed food in front of \pair{him}{.} \\
        \texttt{filler\_gap\_subj} & I know \pair{that}{who} the uncle grabbed food in front of \pair{him}{.} \\
        \bottomrule
    \end{tabular}}
\end{table}
\section{Training and evaluation details}
\label{sec:training_details}

We load models using the HuggingFace \texttt{transformers} \citep{wolf-etal-2020-transformers} library. Up to size \texttt{410m} we load weights in \texttt{float32} precision, \texttt{1b} in \texttt{bfloat16} precision, and larger models in \texttt{float16} precision. Our training set starts with 200 examples sampled according to the scheme in \cref{sec:template}. We then double the size of the set (400) by swapping the base and source inputs/labels and adding these to the training set; including both directions of the intervention makes the comparison fairer between DAS and the other non-paired methods, and also ensures a perfect balance between labels.

The evaluation set consists of 50 examples sampled the same way (effectively 100), except we resample in case we encounter a sentence already present in the training set. Thus, there is no overlap with the training set. We evaluate all metrics (odds-ratio and probe classification accuracy) on this set.

We train DAS for one epoch with a batch size of 4, resulting in 100 backpropagation steps. We use the Adam optimiser  \citep{kingma2014adam} and a linear learning rate schedule, with the first 10\% of training being a warmup from 0 to the learning rate, followed by the learning rate linearly decaying to 0 for the rest of training. The scheduling and optimiser is identical to \citet{wu2023interpretability}. We use a learning rate of $5 \cdot 10^{-3}$, which is higher than previous work (usually $10^{-3}$) due to the small training set size; see \cref{sec:hyperparam} for hyperparameter tuning experiments which justify this choice.

To run our experiments, we used a cluster of NVIDIA A100 (40 GB) and NVIDIA RTX 6000 Ada Generation GPUs. The total runtime for the benchmarking experiments in \cref{sec:experiments} was $\sim 400$ hours, and for the case studies in \cref{sec:studies} it was $\sim 25$ hours.

\section{Hyperparameter tuning}
\label{sec:hyperparam}

To ensure fair comparison, we tuned hyperparameters for DAS, probes, and PCA on a dev set, sampled the same way as the eval set (non-overlapping with train set) but with a different random seed. We train on all tasks in \benchmarktitle{} and report the average odds-ratio following the same evaluation setup as in \cref{sec:efficacious}. We studied only the three smallest models (\texttt{pythia-14m}, \texttt{31m}, \texttt{70m}) due to the large number of experiments needed. Specifically, we tune the learning rate for DAS, the type of regularisation and whether or not to include a bias term in the logit for probes,\footnote{Cf.~\citet{tigges2023linear}, who did not include a bias term in their causal evaluation of probing.} and averaging of the first $c$ components for PCA. We report the overall log odds-ratio for various hyperparameter settings in \cref{tab:params}. 
These experiments were run on a NVIDIA RTX 6000 Ada Generation. The total runtime was $\sim25$ hours.

For probing (\cref{tab:probe_params}), we found that including a bias term and using only $L_2$ regularisation with the \texttt{saga} solver delivers the best performance. However, the setting of the weight coefficient $\lambda$ on the regularisation term in the loss depends on the model. The main architectural difference between these three models is the hidden dimension size, so we suspect that the optimal choice for $\lambda$ depends on that. Roughly extrapolating the observed trend, in our main experiments we check $\lambda = \{10^4, 10^5\}$ for \texttt{pythia-160m} and \texttt{410m}, $\lambda = \{10^5, 10^6\}$ for \texttt{pythia-1b}, \texttt{1.4b}, and \texttt{2.8b}, and $\lambda = \{10^6, 10^7\}$ for \texttt{pythia-6.9b}. As for why $L_2$ regularisation increases causal efficacy, we note that \citet{hewitt-liang-2019-designing} found that it also increases probe selectivity---we leave this as an open question for future work.

For PCA (\cref{tab:pca_params}), we found that averaging the first $c$ components did not improve performance over just using the first component; thus, we used just the first PCA component in our main-text experiments.

For DAS (\cref{tab:das_params}), we found that using the learning rate suggested by \citet{wu2023interpretability}, $10^{-3}$, understated performance and a higher learning rate did not result in any apparent training instability. However, our experimental setup is quite different (smaller training set, no learned boundary, greater variety of model scales). We did not find any consistent differences or trends with model scale between learning rates of $5 \cdot 10^{-3}$ and $10^{-2}$, so we used the former for all experiments.

\begin{table}[!ht]
    \small
    \centering
    \begin{minipage}[c]{0.45\textwidth}
    \begin{subtable}[t]{\textwidth}\centering
    \adjustbox{max width=0.95\textwidth}{\begin{tabular}{llrrrrr}
    \toprule
    \multirow{2}{*}{\textbf{Model}} & 
    \multirow{2}{*}{\textbf{Probe}} & \multicolumn{5}{c}{$\lambda$} \\
    \cmidrule{3-7}
    & & $10^0$ & $10^{1}$ & $10^{2}$ & $10^{3}$ & $10^{4}$ \\
    \midrule
    \textbf{\texttt{pythia-14m}} & No reg., no int. & 0.80 \\
    ($d=128$) & No reg., int. & 0.85 \\
    & $L_1$, no int. & 0.38 & 0.21 & & 0.00 \\
    & $L_1$, int. & 0.41 & 0.22 & & 0.00 \\
    & $L_2$, no int. & 1.07 & 1.15 & & 1.08 \\
    & $L_2$, int. & 1.09 & \textbf{1.18} & 1.15 & 1.07 & 1.05 \\
    & $L_1 + L_2$, no int. & 0.93 & 0.55 & & 0.08 \\
    & $L_1 + L_2$, int. & 0.89 & 0.57 & & 0.08 \\
    \midrule
    \textbf{\texttt{pythia-31m}} & No reg., no int. & 1.75 \\
    ($d=256$) & No reg., int. & 1.77 \\
    & $L_1$, no int. & 0.83 & 0.39 & & 0.12 \\
    & $L_1$, int. & 0.83 & 0.40 & & 0.11 \\
    & $L_2$, no int. & 1.98 & 2.14 & & 2.18 \\
    & $L_2$, int. & 2.03 & 2.22 & \textbf{2.26} & 2.11 & 1.90 \\
    & $L_1 + L_2$, no int. & 1.45 & 0.99 & & 0.14 \\
    & $L_1 + L_2$, int. & 1.42 & 0.93 & & 0.14 \\
    \midrule
    \textbf{\texttt{pythia-70m}} & No reg., no int. & 1.72 \\
    ($d=512$) & No reg., int. & 1.72 \\
    & $L_1$, no int. & 0.74 & 0.32 & & 0.31 \\
    & $L_1$, int. & 0.75 & 0.33 & & 0.17 \\
    & $L_2$, no int. & 1.85 & 2.05 & & 2.43 \\
    & $L_2$, int. & 1.87 & 2.08 & 2.38 & \textbf{2.70} & 2.57 \\
    & $L_1 + L_2$, no int. & 1.11 & 0.67 & & 0.32 \\
    & $L_1 + L_2$, int. & 1.12 & 0.71 & & 0.18 \\
    \bottomrule
    \end{tabular}}
    \caption{Overall odds-ratio across various hyperparameter settings for probes. `Int.' means whether the probe logit has a bias term.}
    \label{tab:probe_params}
    \end{subtable}
    \end{minipage}
    \hspace{1em}
    \begin{minipage}[c]{0.45\textwidth}
    \begin{subtable}[t]{\textwidth}
    \centering
    \adjustbox{max width=\textwidth}{
    \begin{tabular}{lrrrrr}
    \toprule
    
    \multirow{2}{*}{\textbf{Model}} & \multicolumn{5}{c}{\textbf{$c$ (\# components)}}\\
    \cmidrule{2-6}
    & $1$ & $2$ & $3$ & $4$ & $5$ \\
    \midrule
    \textbf{\texttt{pythia-14m}} & \textbf{0.48} & 0.44 & 0.34 & 0.29 & 0.28 \\
    \textbf{\texttt{pythia-31m}} & \textbf{0.86} & 0.82 & 0.59 & 0.49 & 0.43 \\
    \textbf{\texttt{pythia-70m}} & \textbf{1.18} & 0.91 & 0.78 & 0.75 & 0.64 \\
    \bottomrule
    \end{tabular}}
    \caption{Overall odds-ratio across variants of PCA, averaging the first $c$ components.}
    \label{tab:pca_params}
    \end{subtable}
    \begin{subtable}[t]{\textwidth}
    \centering
    \adjustbox{max width=\textwidth}{
    \begin{tabular}{lrrrrrr}
    \\[0.1in]
    \toprule
    \multirow{2}{*}{\textbf{Model}} & \multirow{2}{*}{\textbf{LR}} & \multicolumn{5}{c}{\textbf{Step}} \\
    \cmidrule{3-7}
    & & $0$ & $25$ & $50$ & $75$ & $99$ \\
    \midrule
    \textbf{\texttt{pythia-14m}} & $10^{-3}$ & 0.06 & 0.37 & 1.01 & 1.48 & 1.63 \\
    & $5 \cdot 10^{-3}$ & 0.04 & 2.53 & 3.58 & 3.82 & 3.91 \\
    & $10^{-2}$ & 0.04 & 3.17 & 3.72 & 3.95 & \textbf{4.02} \\
    \midrule
    \textbf{\texttt{pythia-31m}} & $10^{-3}$ & 0.04 & 1.09 & 2.83 & 3.64 & 3.83 \\
    & $5 \cdot 10^{-3}$ & 0.04 & 5.19 & 5.78 & 6.00 & \textbf{6.04} \\
    & $10^{-2}$ & 0.03 & 5.05 & 5.44 & 5.77 & 5.87 \\
    \midrule
    \textbf{\texttt{pythia-70m}} & $10^{-3}$ & 0.02 & 2.25 & 4.69 & 5.42 & 5.57 \\
    & $5 \cdot 10^{-3}$ & 0.02 & 7.21 & 7.48 & 7.54 & 7.55 \\
    & $10^{-2}$ & 0.03 & 6.92 & 7.37 & 7.66 & \textbf{7.75} \\
    \bottomrule
    \end{tabular}}
    \caption{Overall odds-ratio across various learning rates for DAS.}
    \label{tab:das_params}
    \end{subtable}
    \end{minipage}
    \caption{Hyperparameter search results.}
    \label{tab:params}
\end{table}

\section{Data and licensing}
We use the original test suites from SyntaxGym which were described in \citet{hu-etal-2020-systematic}. These were released under the MIT License, and our data release will also use the MIT license for compatibility.

\newpage
\section{Detailed odds-ratio results}
\label{sec:overall_odds}

In these comprehensive results, we include an additional method: vanilla interchange intervention. Instead of as in \cref{eq:dii}, vanilla intervention defines $f^*$ as
\begin{equation}
    f^*_{\text{vanilla}}(\mathbf{b}, \mathbf{s}) = f(\mathbf{s})
\end{equation}
i.e.~it entirely replaces the activation with that of the source input. This is equivalent to $n$-dimensional DII where $f(\mathbf{s}) \in \mathbb{R}^n$, and is a significantly more expressive intervention than any methods we tested.

\subsection{Per-layer}
\label{sec:overall_odds_layer}
\begin{figure}[!h]
    \centering
    \includegraphics[width=\columnwidth]{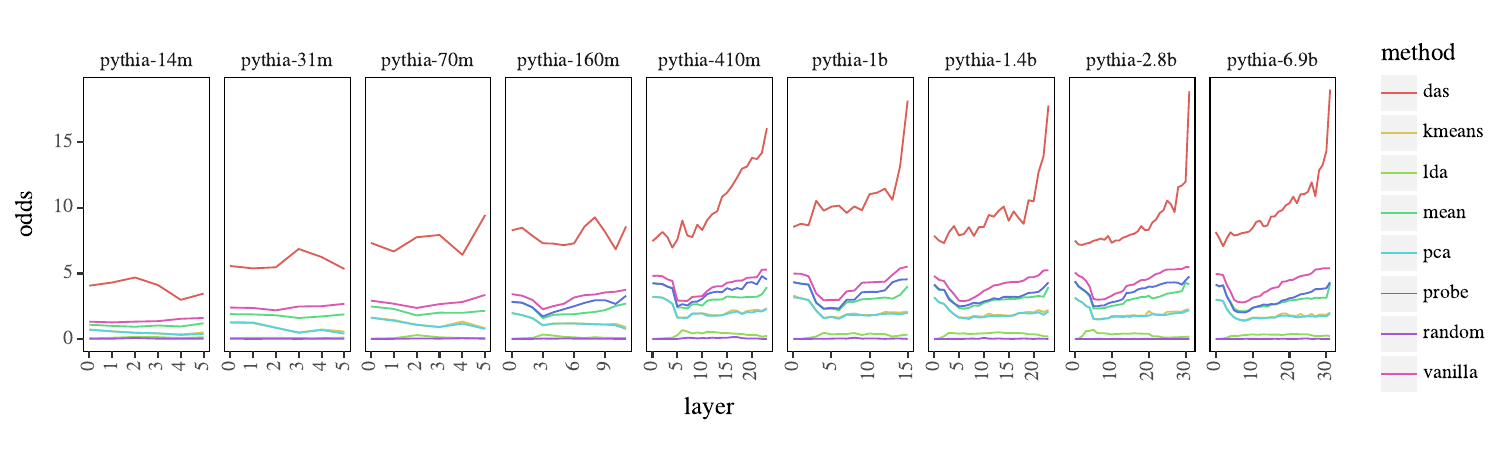}
    \caption{Average odds-ratio per layer and model across all tasks in \benchmarktitle{}.}
    \label{fig:layer-odds}
\end{figure}

\newpage
\subsection{Per-task}
\label{sec:overall_odds_task}
Rows in gray indicate tasks where the model achieves $<60\%$ accuracy.

\begin{table*}[!hp]
    \small
    \centering
    \adjustbox{max width=0.8\textwidth}{
\begin{tabular}{lrrrrrrrrr}
        \toprule
        \multirow{2}{*}{\textbf{Task}} & \multirow{2}{*}{\textbf{Task Acc.}} & \multicolumn{7}{c}{\textbf{Feature-finding methods}} & \multirow{2}{*}{\textbf{Vanilla}} \\
        \cmidrule{3-9}
        & & \textbf{DAS} & \textbf{Probe} & \textbf{Mean} & \textbf{PCA} & \textbf{$k$-means} & \textbf{LDA} & \textbf{Rand.} & \\
        \midrule
\rowcolor{Gainsboro!60}\texttt{agr\_gender} & 0.58 & 0.32 & \textbf{0.94} & 0.49 & 0.35 & 0.36 & 0.03 & 0.01 & 0.81 \\
\texttt{agr\_sv\_num\_subj-relc} & 0.61 & \textbf{2.50} & 1.74 & 1.48 & 0.14 & 0.08 & 0.14 & 0.01 & 1.91 \\
\texttt{agr\_sv\_num\_obj-relc} & 0.79 & 2.03 & 2.02 & 2.05 & 0.30 & 0.32 & 0.06 & 0.03 & \textbf{2.13} \\
\texttt{agr\_sv\_num\_pp} & 0.77 & \textbf{3.47} & 3.15 & 2.85 & 0.36 & 0.14 & 0.15 & 0.03 & 3.28 \\
\texttt{agr\_refl\_num\_subj-relc} & 0.78 & \textbf{2.39} & 2.18 & 1.79 & 0.17 & 0.13 & 0.12 & 0.03 & 2.34 \\
\texttt{agr\_refl\_num\_obj-relc} & 0.72 & \textbf{1.79} & 1.46 & 1.18 & 0.12 & 0.10 & 0.06 & 0.03 & 1.56 \\
\bottomrule
    \end{tabular}}
    \caption{\texttt{pythia-14m}}
    \label{tab:odds_14m}
\end{table*}

\begin{table*}[!hp]
    \small
    \centering
    \adjustbox{max width=0.8\textwidth}{
    \begin{tabular}{lrrrrrrrrr}
        \toprule
        \multirow{2}{*}{\textbf{Task}} & \multirow{2}{*}{\textbf{Task Acc.}} & \multicolumn{7}{c}{\textbf{Feature-finding methods}} & \multirow{2}{*}{\textbf{Vanilla}} \\
        \cmidrule{3-9}
        & & \textbf{DAS} & \textbf{Probe} & \textbf{Mean} & \textbf{PCA} & \textbf{$k$-means} & \textbf{LDA} & \textbf{Rand.} & \\
        \midrule
\rowcolor{Gainsboro!60}\texttt{agr\_gender} & 0.58 & 0.07 & \textbf{1.01} & 0.44 & 0.26 & 0.28 & 0.04 & 0.05 & 0.78 \\
\texttt{agr\_sv\_num\_subj-relc} & 0.61 & 2.03 & 2.65 & 2.23 & 0.20 & 0.12 & 0.19 & 0.02 & \textbf{2.76} \\
\texttt{agr\_sv\_num\_obj-relc} & 0.79 & 1.05 & 2.87 & 2.94 & 0.32 & 0.37 & 0.11 & 0.03 & \textbf{3.03} \\
\texttt{agr\_sv\_num\_pp} & 0.77 & 2.62 & 4.66 & 4.20 & 0.55 & 0.22 & 0.22 & 0.04 & \textbf{4.82} \\
\texttt{agr\_refl\_num\_subj-relc} & 0.78 & 1.97 & 2.93 & 2.39 & 0.17 & 0.12 & 0.14 & 0.05 & \textbf{3.05} \\
\texttt{agr\_refl\_num\_obj-relc} & 0.72 & 1.14 & 1.86 & 1.44 & 0.11 & 0.11 & 0.09 & 0.05 & \textbf{1.95} \\
\bottomrule
    \end{tabular}}
    \caption{\texttt{pythia-14m} (selectivity)}
    \label{tab:odds_adj_14m}
\end{table*}

\begin{table*}[!hp]
    \small
    \centering
    \adjustbox{max width=0.8\textwidth}{
    \begin{tabular}{lrrrrrrrrr}
        \toprule
        \multirow{2}{*}{\textbf{Task}} & \multirow{2}{*}{\textbf{Task Acc.}} & \multicolumn{7}{c}{\textbf{Feature-finding methods}} & \multirow{2}{*}{\textbf{Vanilla}} \\
        \cmidrule{3-9}
        & & \textbf{DAS} & \textbf{Probe} & \textbf{Mean} & \textbf{PCA} & \textbf{$k$-means} & \textbf{LDA} & \textbf{Rand.} & \\
        \midrule
\texttt{agr\_gender} & 0.85 & \textbf{2.50} & 1.74 & 0.52 & 0.13 & 0.12 & 0.01 & 0.03 & 2.04 \\
\texttt{agr\_sv\_num\_subj-relc} & 0.85 & \textbf{4.56} & 3.88 & 2.84 & 0.23 & 0.20 & 0.12 & 0.01 & 3.93 \\
\texttt{agr\_sv\_num\_obj-relc} & 0.94 & \textbf{4.44} & 3.95 & 3.42 & 0.26 & 0.20 & 0.04 & 0.03 & 4.02 \\
\texttt{agr\_sv\_num\_pp} & 0.77 & \textbf{3.62} & 3.28 & 2.37 & 0.32 & 0.18 & 0.05 & 0.02 & 3.31 \\
\texttt{agr\_refl\_num\_subj-relc} & 0.87 & \textbf{3.84} & 3.50 & 2.54 & 0.17 & 0.13 & 0.17 & 0.02 & 3.68 \\
\texttt{agr\_refl\_num\_obj-relc} & 0.88 & 3.66 & 3.47 & 2.57 & 0.20 & 0.17 & 0.15 & 0.03 & \textbf{3.68} \\
\bottomrule
    \end{tabular}}
    \caption{\texttt{pythia-31m}}
    \label{tab:odds_31m}
\end{table*}

\begin{table*}[!hp]
    \small
    \centering
    \adjustbox{max width=0.8\textwidth}{
    \begin{tabular}{lrrrrrrrrr}
        \toprule
        \multirow{2}{*}{\textbf{Task}} & \multirow{2}{*}{\textbf{Task Acc.}} & \multicolumn{7}{c}{\textbf{Feature-finding methods}} & \multirow{2}{*}{\textbf{Vanilla}} \\
        \cmidrule{3-9}
        & & \textbf{DAS} & \textbf{Probe} & \textbf{Mean} & \textbf{PCA} & \textbf{$k$-means} & \textbf{LDA} & \textbf{Rand.} & \\
        \midrule
\texttt{agr\_gender} & 0.85 & \textbf{2.38} & 1.81 & 0.66 & 0.45 & 0.41 & 0.01 & 0.06 & 2.02 \\
\texttt{agr\_sv\_num\_subj-relc} & 0.85 & 4.06 & 5.43 & 4.12 & 0.36 & 0.28 & 0.16 & 0.02 & \textbf{5.63} \\
\texttt{agr\_sv\_num\_obj-relc} & 0.94 & 3.30 & 5.30 & 4.83 & 0.38 & 0.28 & 0.05 & 0.03 & \textbf{5.48} \\
\texttt{agr\_sv\_num\_pp} & 0.77 & 2.87 & 4.80 & 3.48 & 0.49 & 0.28 & 0.09 & 0.03 & \textbf{4.92} \\
\texttt{agr\_refl\_num\_subj-relc} & 0.87 & 2.83 & 4.34 & 3.10 & 0.24 & 0.16 & 0.22 & 0.02 & \textbf{4.48} \\
\texttt{agr\_refl\_num\_obj-relc} & 0.88 & 2.38 & 4.28 & 3.14 & 0.26 & 0.21 & 0.15 & 0.03 & \textbf{4.44} \\
\bottomrule
    \end{tabular}}
    \caption{\texttt{pythia-31m} (selectivity)}
    \label{tab:odds_31m_s}
\end{table*}

\begin{table*}[!hp]
    \small
    \centering
    \adjustbox{max width=0.8\textwidth}{
    \begin{tabular}{lrrrrrrrrr}
        \toprule
        \multirow{2}{*}{\textbf{Task}} & \multirow{2}{*}{\textbf{Task Acc.}} & \multicolumn{7}{c}{\textbf{Feature-finding methods}} & \multirow{2}{*}{\textbf{Vanilla}} \\
        \cmidrule{3-9}
        & & \textbf{DAS} & \textbf{Probe} & \textbf{Mean} & \textbf{PCA} & \textbf{$k$-means} & \textbf{LDA} & \textbf{Rand.} & \\
        \midrule\texttt{agr\_gender} & 0.95 & \textbf{3.29} & 2.95 & 1.09 & 0.74 & 0.77 & 0.06 & 0.04 & 2.99 \\
\texttt{agr\_sv\_num\_subj-relc} & 0.97 & \textbf{4.77} & 4.11 & 3.66 & 0.53 & 0.67 & 0.42 & 0.01 & 4.34 \\
\texttt{agr\_sv\_num\_obj-relc} & 0.86 & \textbf{3.60} & 3.38 & 3.36 & 0.49 & 1.10 & 0.39 & 0.01 & 3.44 \\
\texttt{agr\_sv\_num\_pp} & 1.00 & \textbf{5.83} & 4.66 & 4.20 & 0.47 & 0.74 & 0.21 & 0.01 & 4.96 \\
\texttt{agr\_refl\_num\_subj-relc} & 0.93 & \textbf{5.35} & 3.80 & 2.88 & 0.32 & 0.22 & 0.18 & 0.01 & 3.99 \\
\texttt{agr\_refl\_num\_obj-relc} & 0.90 & \textbf{3.97} & 2.94 & 2.16 & 0.29 & 0.22 & 0.15 & 0.01 & 3.02 \\
\bottomrule
    \end{tabular}}
    \caption{\texttt{pythia-70m}}
    \label{tab:odds_70m}
\end{table*}

\begin{table*}[!hp]
    \small
    \centering
    \adjustbox{max width=0.8\textwidth}{
    \begin{tabular}{lrrrrrrrrr}
        \toprule
        \multirow{2}{*}{\textbf{Task}} & \multirow{2}{*}{\textbf{Task Acc.}} & \multicolumn{7}{c}{\textbf{Feature-finding methods}} & \multirow{2}{*}{\textbf{Vanilla}} \\
        \cmidrule{3-9}
        & & \textbf{DAS} & \textbf{Probe} & \textbf{Mean} & \textbf{PCA} & \textbf{$k$-means} & \textbf{LDA} & \textbf{Rand.} & \\
        \midrule
\texttt{agr\_gender} & 0.95 & 1.68 & 2.75 & 1.54 & 1.19 & 1.25 & 0.06 & 0.04 & \textbf{3.03} \\
\texttt{agr\_sv\_num\_subj-relc} & 0.97 & 3.08 & 5.34 & 4.76 & 0.67 & 0.85 & 0.55 & 0.02 & \textbf{5.69} \\
\texttt{agr\_sv\_num\_obj-relc} & 0.86 & 2.30 & 4.47 & 4.45 & 0.65 & 1.44 & 0.51 & 0.02 & \textbf{4.59} \\
\texttt{agr\_sv\_num\_pp} & 1.00 & 4.19 & 6.11 & 5.49 & 0.60 & 0.97 & 0.30 & 0.02 & \textbf{6.55} \\
\texttt{agr\_refl\_num\_subj-relc} & 0.93 & 3.63 & 4.89 & 3.76 & 0.44 & 0.25 & 0.25 & 0.01 & \textbf{5.04} \\
\texttt{agr\_refl\_num\_obj-relc} & 0.90 & 3.35 & 3.77 & 2.78 & 0.35 & 0.24 & 0.19 & 0.01 & \textbf{3.80} \\
\bottomrule
    \end{tabular}}
    \caption{\texttt{pythia-70m} (selectivity)}
    \label{tab:odds_70m_s}
\end{table*}

\begin{table*}[!hp]
    \small
    \centering
    \adjustbox{max width=0.9\textwidth}{
    \begin{tabular}{lrrrrrrrrrr}
        \toprule
        \multirow{2}{*}{\textbf{Task}} & \multirow{2}{*}{\textbf{Task Acc.}} & \multicolumn{8}{c}{\textbf{Feature-finding methods}} & \multirow{2}{*}{\textbf{Vanilla}} \\
        \cmidrule{3-10}
        & & \textbf{DAS} & \textbf{Probe}$^0$ & \textbf{Probe}$^1$ & \textbf{Mean} & \textbf{PCA} & \textbf{$k$-means} & \textbf{LDA} & \textbf{Rand.} & \\
        \midrule
\texttt{agr\_gender} & 0.99 & \textbf{5.64} & 3.89 & 2.87 & 1.52 & 0.96 & 0.96 & 0.06 & 0.03 & 4.35 \\
\texttt{agr\_sv\_num\_subj-relc} & 0.96 & \textbf{4.79} & 4.10 & 3.20 & 2.52 & 0.19 & 0.22 & 0.26 & 0.01 & 4.20 \\
\texttt{agr\_sv\_num\_obj-relc} & 0.95 & \textbf{4.52} & 4.49 & 4.01 & 3.22 & 0.24 & 0.26 & 0.36 & 0.00 & 4.19 \\
\texttt{agr\_sv\_num\_pp} & 0.98 & \textbf{5.02} & 4.42 & 3.48 & 2.91 & 0.23 & 0.20 & 0.25 & 0.01 & 4.49 \\
\texttt{agr\_refl\_num\_subj-relc} & 0.92 & \textbf{4.56} & 3.93 & 2.85 & 1.99 & 0.11 & 0.13 & 0.28 & 0.01 & 3.91 \\
\texttt{agr\_refl\_num\_obj-relc} & 0.94 & \textbf{4.65} & 3.74 & 2.71 & 1.97 & 0.19 & 0.17 & 0.38 & 0.00 & 3.83 \\
\bottomrule
    \end{tabular}}
    \caption{\texttt{pythia-160m}; Probe$^0$ has $\lambda = 10^4$, Probe$^1$ has $\lambda = 10^5$.}
    \label{tab:odds_160m}
\end{table*}

\begin{table*}[!hp]
    \small
    \centering
    \adjustbox{max width=0.9\textwidth}{
    \begin{tabular}{lrrrrrrrrrr}
        \toprule
        \multirow{2}{*}{\textbf{Task}} & \multirow{2}{*}{\textbf{Task Acc.}} & \multicolumn{8}{c}{\textbf{Feature-finding methods}} & \multirow{2}{*}{\textbf{Vanilla}} \\
        \cmidrule{3-10}
        & & \textbf{DAS} & \textbf{Probe}$^0$ & \textbf{Probe}$^1$ & \textbf{Mean} & \textbf{PCA} & \textbf{$k$-means} & \textbf{LDA} & \textbf{Rand.} & \\
        \midrule\texttt{agr\_gender} & 0.99 & \textbf{4.85} & 4.30 & 3.33 & 1.72 & 1.17 & 1.19 & 0.06 & 0.05 & 4.49 \\
\texttt{agr\_sv\_num\_subj-relc} & 0.96 & 3.66 & 5.37 & 4.21 & 3.33 & 0.29 & 0.31 & 0.32 & 0.02 & \textbf{5.47} \\
\texttt{agr\_sv\_num\_obj-relc} & 0.95 & 3.65 & \textbf{6.09} & 5.45 & 4.37 & 0.41 & 0.44 & 0.47 & 0.01 & 5.71 \\
\texttt{agr\_sv\_num\_pp} & 0.98 & 3.58 & \textbf{5.94} & 4.67 & 3.88 & 0.34 & 0.32 & 0.33 & 0.02 & 5.88 \\
\texttt{agr\_refl\_num\_subj-relc} & 0.92 & 3.44 & \textbf{4.43} & 3.21 & 2.21 & 0.12 & 0.14 & 0.32 & 0.01 & 4.25 \\
\texttt{agr\_refl\_num\_obj-relc} & 0.94 & 3.25 & \textbf{4.30} & 3.11 & 2.23 & 0.22 & 0.21 & 0.43 & 0.00 & 4.29 \\
\bottomrule
    \end{tabular}}
    \caption{\texttt{pythia-160m} (selectivity)}
    \label{tab:odds_160m_s}
\end{table*}

\begin{table*}[!hp]
    \small
    \centering
    \adjustbox{max width=0.9\textwidth}{
    \begin{tabular}{lrrrrrrrrrr}
        \toprule
        \multirow{2}{*}{\textbf{Task}} & \multirow{2}{*}{\textbf{Task Acc.}} & \multicolumn{8}{c}{\textbf{Feature-finding methods}} & \multirow{2}{*}{\textbf{Vanilla}} \\
        \cmidrule{3-10}
        & & \textbf{DAS} & \textbf{Probe}$^0$ & \textbf{Probe}$^1$ & \textbf{Mean} & \textbf{PCA} & \textbf{$k$-means} & \textbf{LDA} & \textbf{Rand.} & \\
        \midrule\texttt{agr\_gender} & 1.00 & 3.86 & 2.99 & 2.80 & 1.45 & 1.03 & 1.03 & 0.05 & 0.06 & \textbf{4.01} \\
\texttt{agr\_sv\_num\_subj-relc} & 0.97 & \textbf{4.97} & 4.32 & 4.19 & 3.72 & 0.29 & 0.28 & 1.88 & 0.03 & 4.42 \\
\texttt{agr\_sv\_num\_obj-relc} & 0.99 & \textbf{5.61} & 5.01 & 5.46 & 4.52 & 0.32 & 0.29 & 2.22 & 0.03 & 5.01 \\
\texttt{agr\_sv\_num\_pp} & 0.97 & \textbf{5.63} & 4.96 & 4.83 & 4.55 & 0.38 & 0.35 & 1.15 & 0.04 & 5.09 \\
\texttt{agr\_refl\_num\_subj-relc} & 0.92 & 3.65 & 3.86 & 3.77 & 1.85 & 0.15 & 0.14 & 0.61 & 0.02 & \textbf{3.90} \\
\texttt{agr\_refl\_num\_obj-relc} & 0.96 & \textbf{4.43} & 4.03 & 4.04 & 1.90 & 0.31 & 0.31 & 0.63 & 0.02 & 3.96 \\
\bottomrule
    \end{tabular}}
    \caption{\texttt{pythia-410m}; Probe$^0$ has $\lambda = 10^4$, Probe$^1$ has $\lambda = 10^5$.}
    \label{tab:odds_410m}
\end{table*}

\begin{table*}[!hp]
    \small
    \centering
    \adjustbox{max width=0.9\textwidth}{
    \begin{tabular}{lrrrrrrrrrr}
        \toprule
        \multirow{2}{*}{\textbf{Task}} & \multirow{2}{*}{\textbf{Task Acc.}} & \multicolumn{8}{c}{\textbf{Feature-finding methods}} & \multirow{2}{*}{\textbf{Vanilla}} \\
        \cmidrule{3-10}
        & & \textbf{DAS} & \textbf{Probe}$^0$ & \textbf{Probe}$^1$ & \textbf{Mean} & \textbf{PCA} & \textbf{$k$-means} & \textbf{LDA} & \textbf{Rand.} & \\
        \midrule
\texttt{agr\_gender} & 1.00 & 2.77 & 3.06 & 3.00 & 1.64 & 1.29 & 1.29 & 0.05 & 0.12 & \textbf{4.08} \\
\texttt{agr\_sv\_num\_subj-relc} & 0.97 & 3.64 & \textbf{5.63} & 5.45 & 4.84 & 0.40 & 0.38 & 2.43 & 0.03 & 5.52 \\
\texttt{agr\_sv\_num\_obj-relc} & 0.99 & 4.05 & 6.86 & \textbf{7.42} & 6.01 & 0.42 & 0.38 & 2.93 & 0.05 & 6.78 \\
\texttt{agr\_sv\_num\_pp} & 0.97 & 4.46 & 6.51 & 6.32 & 5.95 & 0.50 & 0.47 & 1.49 & 0.05 & \textbf{6.55} \\
\texttt{agr\_refl\_num\_subj-relc} & 0.92 & 2.79 & \textbf{4.42} & 4.23 & 2.10 & 0.15 & 0.11 & 0.72 & 0.02 & 4.36 \\
\texttt{agr\_refl\_num\_obj-relc} & 0.96 & 2.76 & \textbf{4.79} & 4.78 & 2.26 & 0.34 & 0.33 & 0.81 & 0.02 & 4.61 \\
\bottomrule
    \end{tabular}}
    \caption{\texttt{pythia-410m} (selectivity)}
    \label{tab:odds_410m_s}
\end{table*}

\begin{table*}[!hp]
    \small
    \centering
    \adjustbox{max width=0.9\textwidth}{
    \begin{tabular}{lrrrrrrrrrr}
        \toprule
        \multirow{2}{*}{\textbf{Task}} & \multirow{2}{*}{\textbf{Task Acc.}} & \multicolumn{8}{c}{\textbf{Feature-finding methods}} & \multirow{2}{*}{\textbf{Vanilla}} \\
        \cmidrule{3-10}
        & & \textbf{DAS} & \textbf{Probe}$^0$ & \textbf{Probe}$^1$ & \textbf{Mean} & \textbf{PCA} & \textbf{$k$-means} & \textbf{LDA} & \textbf{Rand.} & \\
        \midrule\texttt{agr\_gender} & 1.00 & \textbf{4.72} & 2.43 & 2.27 & 1.32 & 0.90 & 0.90 & 0.01 & 0.05 & 3.85 \\
\texttt{agr\_sv\_num\_subj-relc} & 1.00 & \textbf{6.62} & 5.23 & 4.98 & 4.24 & 0.36 & 0.39 & 1.40 & 0.01 & 5.51 \\
\texttt{agr\_sv\_num\_obj-relc} & 0.94 & \textbf{5.30} & 4.75 & 4.62 & 3.72 & 0.38 & 0.41 & 1.39 & 0.02 & 4.44 \\
\texttt{agr\_sv\_num\_pp} & 0.94 & \textbf{5.81} & 4.78 & 4.45 & 3.86 & 0.36 & 0.40 & 0.82 & 0.03 & 4.91 \\
\texttt{agr\_refl\_num\_subj-relc} & 0.85 & \textbf{4.13} & 3.57 & 2.76 & 1.78 & 0.17 & 0.21 & 0.48 & 0.01 & 3.51 \\
\texttt{agr\_refl\_num\_obj-relc} & 1.00 & \textbf{5.89} & 4.71 & 4.09 & 2.60 & 0.39 & 0.42 & 0.47 & 0.00 & 4.86 \\
\bottomrule
    \end{tabular}}
    \caption{\texttt{pythia-1b}; Probe$^0$ has $\lambda = 10^5$, Probe$^1$ has $\lambda = 10^6$.}
    \label{tab:odds_1b}
\end{table*}

\begin{table*}[!hp]
    \small
    \centering
    \adjustbox{max width=0.9\textwidth}{
    \begin{tabular}{lrrrrrrrrrr}
        \toprule
        \multirow{2}{*}{\textbf{Task}} & \multirow{2}{*}{\textbf{Task Acc.}} & \multicolumn{8}{c}{\textbf{Feature-finding methods}} & \multirow{2}{*}{\textbf{Vanilla}} \\
        \cmidrule{3-10}
        & & \textbf{DAS} & \textbf{Probe}$^0$ & \textbf{Probe}$^1$ & \textbf{Mean} & \textbf{PCA} & \textbf{$k$-means} & \textbf{LDA} & \textbf{Rand.} & \\
        \midrule
\texttt{agr\_gender} & 1.00 & 3.66 & 3.00 & 2.73 & 2.06 & 1.65 & 1.65 & 0.01 & 0.09 & \textbf{4.33} \\
\texttt{agr\_sv\_num\_subj-relc} & 1.00 & 4.13 & 6.97 & 6.59 & 5.60 & 0.42 & 0.46 & 1.75 & 0.01 & \textbf{7.32} \\
\texttt{agr\_sv\_num\_obj-relc} & 0.94 & 3.94 & \textbf{6.55} & 6.32 & 5.10 & 0.41 & 0.45 & 1.82 & 0.03 & 6.01 \\
\texttt{agr\_sv\_num\_pp} & 0.94 & 4.74 & 6.26 & 5.78 & 5.00 & 0.47 & 0.51 & 1.01 & 0.05 & \textbf{6.27} \\
\texttt{agr\_refl\_num\_subj-relc} & 0.85 & 3.20 & \textbf{4.13} & 3.19 & 2.09 & 0.12 & 0.15 & 0.62 & 0.01 & 4.10 \\
\texttt{agr\_refl\_num\_obj-relc} & 1.00 & 3.49 & 5.21 & 4.50 & 2.83 & 0.31 & 0.37 & 0.59 & 0.01 & \textbf{5.26} \\
\bottomrule
    \end{tabular}}
    \caption{\texttt{pythia-1b} (selectivity)}
    \label{tab:odds_1b_s}
\end{table*}

\begin{table*}[!hp]
    \small
    \centering
    \adjustbox{max width=0.9\textwidth}{
    \begin{tabular}{lrrrrrrrrrr}
        \toprule
        \multirow{2}{*}{\textbf{Task}} & \multirow{2}{*}{\textbf{Task Acc.}} & \multicolumn{8}{c}{\textbf{Feature-finding methods}} & \multirow{2}{*}{\textbf{Vanilla}} \\
        \cmidrule{3-10}
        & & \textbf{DAS} & \textbf{Probe}$^0$ & \textbf{Probe}$^1$ & \textbf{Mean} & \textbf{PCA} & \textbf{$k$-means} & \textbf{LDA} & \textbf{Rand.} & \\
        \midrule\texttt{agr\_gender} & 1.00 & \textbf{3.62} & 2.58 & 2.00 & 1.11 & 0.58 & 0.58 & 0.00 & 0.02 & 3.24 \\
\texttt{agr\_sv\_num\_subj-relc} & 0.98 & \textbf{4.82} & 4.24 & 4.07 & 3.89 & 0.48 & 0.44 & 2.43 & 0.02 & 4.32 \\
\texttt{agr\_sv\_num\_obj-relc} & 0.97 & \textbf{5.11} & 4.89 & 4.46 & 3.97 & 0.49 & 0.61 & 2.37 & 0.02 & 4.44 \\
\texttt{agr\_sv\_num\_pp} & 0.99 & \textbf{5.75} & 4.94 & 4.77 & 4.58 & 0.54 & 0.37 & 0.71 & 0.01 & 5.11 \\
\texttt{agr\_refl\_num\_subj-relc} & 0.94 & \textbf{3.44} & 3.27 & 2.19 & 1.83 & 0.17 & 0.38 & 0.92 & 0.01 & 3.31 \\
\texttt{agr\_refl\_num\_obj-relc} & 0.99 & \textbf{4.02} & 3.96 & 2.71 & 1.98 & 0.29 & 0.39 & 0.91 & 0.01 & 3.85 \\
\bottomrule
    \end{tabular}}
    \caption{\texttt{pythia-1.4b}; Probe$^0$ has $\lambda = 10^5$, Probe$^1$ has $\lambda = 10^6$.}
    \label{tab:odds_1.4b}
\end{table*}

\begin{table*}[!hp]
    \small
    \centering
    \adjustbox{max width=0.9\textwidth}{
    \begin{tabular}{lrrrrrrrrrr}
        \toprule
        \multirow{2}{*}{\textbf{Task}} & \multirow{2}{*}{\textbf{Task Acc.}} & \multicolumn{8}{c}{\textbf{Feature-finding methods}} & \multirow{2}{*}{\textbf{Vanilla}} \\
        \cmidrule{3-10}
        & & \textbf{DAS} & \textbf{Probe}$^0$ & \textbf{Probe}$^1$ & \textbf{Mean} & \textbf{PCA} & \textbf{$k$-means} & \textbf{LDA} & \textbf{Rand.} & \\
        \midrule\texttt{agr\_gender} & 1.00 & 2.59 & 2.66 & 2.18 & 1.29 & 0.86 & 0.86 & 0.00 & 0.04 & \textbf{3.33} \\
\texttt{agr\_sv\_num\_subj-relc} & 0.98 & 3.52 & \textbf{5.70} & 5.43 & 5.19 & 0.59 & 0.56 & 3.20 & 0.02 & 5.67 \\
\texttt{agr\_sv\_num\_obj-relc} & 0.97 & 3.36 & \textbf{6.73} & 6.08 & 5.21 & 0.63 & 0.79 & 3.06 & 0.03 & 6.08 \\
\texttt{agr\_sv\_num\_pp} & 0.99 & 4.14 & 6.71 & 6.42 & 6.15 & 0.72 & 0.52 & 0.87 & 0.03 & \textbf{6.85} \\
\texttt{agr\_refl\_num\_subj-relc} & 0.94 & 2.76 & 3.86 & 2.63 & 2.32 & 0.20 & 0.50 & 1.24 & 0.01 & \textbf{3.89} \\
\texttt{agr\_refl\_num\_obj-relc} & 0.99 & 2.91 & \textbf{4.52} & 3.05 & 2.25 & 0.25 & 0.36 & 1.06 & 0.01 & 4.31 \\
\bottomrule
    \end{tabular}}
    \caption{\texttt{pythia-1.4b} (selectivity)}
    \label{tab:odds_1.4b_s}
\end{table*}

\begin{table*}[!hp]
    \small
    \centering
    \adjustbox{max width=0.9\textwidth}{
    \begin{tabular}{lrrrrrrrrrr}
        \toprule
        \multirow{2}{*}{\textbf{Task}} & \multirow{2}{*}{\textbf{Task Acc.}} & \multicolumn{8}{c}{\textbf{Feature-finding methods}} & \multirow{2}{*}{\textbf{Vanilla}} \\
        \cmidrule{3-10}
        & & \textbf{DAS} & \textbf{Probe}$^0$ & \textbf{Probe}$^1$ & \textbf{Mean} & \textbf{PCA} & \textbf{$k$-means} & \textbf{LDA} & \textbf{Rand.} & \\
        \midrule
\texttt{agr\_gender} & 1.00 & \textbf{4.99} & 2.53 & 1.89 & 1.08 & 0.38 & 0.38 & 0.01 & 0.01 & 3.54 \\
\texttt{agr\_sv\_num\_subj-relc} & 0.97 & \textbf{5.87} & 5.15 & 4.99 & 4.19 & 0.34 & 0.35 & 2.21 & 0.00 & 5.30 \\
\texttt{agr\_sv\_num\_obj-relc} & 0.97 & \textbf{6.14} & 5.68 & 5.88 & 5.06 & 0.39 & 0.38 & 2.28 & 0.00 & 5.65 \\
\texttt{agr\_sv\_num\_pp} & 0.97 & \textbf{6.21} & 5.69 & 5.38 & 4.64 & 0.34 & 0.33 & 0.27 & 0.00 & 5.92 \\
\texttt{agr\_refl\_num\_subj-relc} & 0.97 & \textbf{4.90} & 3.48 & 2.89 & 2.06 & 0.16 & 0.16 & 0.94 & 0.00 & 3.94 \\
\texttt{agr\_refl\_num\_obj-relc} & 0.97 & \textbf{5.98} & 4.15 & 3.32 & 1.95 & 0.31 & 0.34 & 0.76 & 0.01 & 4.66 \\
\bottomrule
    \end{tabular}}
    \caption{\texttt{pythia-2.8b}; Probe$^0$ has $\lambda = 10^5$, Probe$^1$ has $\lambda = 10^6$.}
    \label{tab:odds_2.8b}
\end{table*}

\begin{table*}[!hp]
    \small
    \centering
    \adjustbox{max width=0.9\textwidth}{
    \begin{tabular}{lrrrrrrrrrr}
        \toprule
        \multirow{2}{*}{\textbf{Task}} & \multirow{2}{*}{\textbf{Task Acc.}} & \multicolumn{8}{c}{\textbf{Feature-finding methods}} & \multirow{2}{*}{\textbf{Vanilla}} \\
        \cmidrule{3-10}
        & & \textbf{DAS} & \textbf{Probe}$^0$ & \textbf{Probe}$^1$ & \textbf{Mean} & \textbf{PCA} & \textbf{$k$-means} & \textbf{LDA} & \textbf{Rand.} & \\
        \midrule
\texttt{agr\_gender} & 1.00 & \textbf{4.21} & 2.63 & 2.36 & 1.62 & 1.06 & 1.07 & 0.01 & 0.02 & 3.72 \\
\texttt{agr\_sv\_num\_subj-relc} & 0.97 & 3.13 & 6.91 & 6.65 & 5.57 & 0.41 & 0.42 & 2.92 & 0.01 & \textbf{7.07} \\
\texttt{agr\_sv\_num\_obj-relc} & 0.97 & 3.47 & 7.60 & \textbf{7.79} & 6.69 & 0.41 & 0.38 & 2.98 & 0.01 & 7.49 \\
\texttt{agr\_sv\_num\_pp} & 0.97 & 4.61 & 7.94 & 7.49 & 6.42 & 0.48 & 0.48 & 0.37 & 0.01 & \textbf{8.00} \\
\texttt{agr\_refl\_num\_subj-relc} & 0.97 & 3.18 & 4.21 & 3.42 & 2.45 & 0.15 & 0.14 & 1.15 & 0.00 & \textbf{4.62} \\
\texttt{agr\_refl\_num\_obj-relc} & 0.97 & 2.62 & 4.70 & 3.71 & 2.19 & 0.24 & 0.30 & 0.90 & 0.01 & \textbf{5.06} \\
\bottomrule
    \end{tabular}}
    \caption{\texttt{pythia-2.8b} (selectivity)}
    \label{tab:odds_2.8b_s}
\end{table*}

\begin{table*}[!hp]
    \small
    \centering
    \adjustbox{max width=0.9\textwidth}{
    \begin{tabular}{lrrrrrrrrrr}
        \toprule
        \multirow{2}{*}{\textbf{Task}} & \multirow{2}{*}{\textbf{Task Acc.}} & \multicolumn{8}{c}{\textbf{Feature-finding methods}} & \multirow{2}{*}{\textbf{Vanilla}} \\
        \cmidrule{3-10}
        & & \textbf{DAS} & \textbf{Probe}$^0$ & \textbf{Probe}$^1$ & \textbf{Mean} & \textbf{PCA} & \textbf{$k$-means} & \textbf{LDA} & \textbf{Rand.} & \\
        \midrule
\texttt{agr\_gender} & 0.99 & \textbf{4.18} & 3.58 & 2.36 & 1.37 & 0.56 & 0.56 & 0.05 & 0.01 & 3.81 \\
\texttt{agr\_sv\_num\_subj-relc} & 0.99 & \textbf{5.23} & 4.42 & 3.86 & 3.56 & 0.32 & 0.30 & 2.22 & 0.01 & 4.68 \\
\texttt{agr\_sv\_num\_obj-relc} & 0.99 & 5.82 & \textbf{6.16} & 5.23 & 4.31 & 0.31 & 0.27 & 2.10 & 0.01 & 5.55 \\
\texttt{agr\_sv\_num\_pp} & 0.99 & \textbf{4.93} & 4.16 & 3.66 & 3.44 & 0.32 & 0.28 & 0.04 & 0.01 & 4.56 \\
\texttt{agr\_refl\_num\_subj-relc} & 0.94 & 4.00 & 3.52 & 2.41 & 2.24 & 0.14 & 0.13 & 0.07 & 0.01 & \textbf{4.12} \\
\texttt{agr\_refl\_num\_obj-relc} & 1.00 & \textbf{5.06} & 4.48 & 2.86 & 2.23 & 0.25 & 0.27 & 0.09 & 0.01 & 4.58 \\
\bottomrule
    \end{tabular}}
    \caption{\texttt{pythia-6.9b}; Probe$^0$ has $\lambda = 10^6$, Probe$^1$ has $\lambda = 10^7$.}
    \label{tab:odds_6.9b}
\end{table*}

\begin{table*}[!hp]
    \small
    \centering
    \adjustbox{max width=0.9\textwidth}{
    \begin{tabular}{lrrrrrrrrrr}
        \toprule
        \multirow{2}{*}{\textbf{Task}} & \multirow{2}{*}{\textbf{Task Acc.}} & \multicolumn{8}{c}{\textbf{Feature-finding methods}} & \multirow{2}{*}{\textbf{Vanilla}} \\
        \cmidrule{3-10}
        & & \textbf{DAS} & \textbf{Probe}$^0$ & \textbf{Probe}$^1$ & \textbf{Mean} & \textbf{PCA} & \textbf{$k$-means} & \textbf{LDA} & \textbf{Rand.} & \\
        \midrule
\texttt{agr\_gender} & 0.99 & 3.15 & 4.08 & 2.82 & 1.99 & 1.07 & 1.07 & 0.04 & 0.02 & \textbf{4.20} \\
\texttt{agr\_sv\_num\_subj-relc} & 0.99 & 3.84 & 5.91 & 5.13 & 4.71 & 0.42 & 0.38 & 2.86 & 0.02 & \textbf{6.34} \\
\texttt{agr\_sv\_num\_obj-relc} & 0.99 & 4.40 & \textbf{8.22} & 6.95 & 5.67 & 0.33 & 0.29 & 2.74 & 0.02 & 7.42 \\
\texttt{agr\_sv\_num\_pp} & 0.99 & 3.58 & 5.83 & 5.09 & 4.76 & 0.42 & 0.36 & 0.05 & 0.02 & \textbf{6.28} \\
\texttt{agr\_refl\_num\_subj-relc} & 0.94 & 2.28 & 4.10 & 2.82 & 2.59 & 0.14 & 0.10 & 0.09 & 0.01 & \textbf{4.65} \\
\texttt{agr\_refl\_num\_obj-relc} & 1.00 & 2.74 & 5.03 & 3.19 & 2.49 & 0.19 & 0.23 & 0.11 & 0.01 & \textbf{5.12} \\
\bottomrule
    \end{tabular}}
    \caption{\texttt{pythia-6.9b} (selectivity)}
    \label{tab:odds_6.9b_s}
\end{table*}

\newpage
\section{Odds-ratio plots for all methods on selected tasks}
\label{sec:odds_plots}
\begin{figure*}[!hp]
    \centering
    \includegraphics[width=0.85\textwidth]{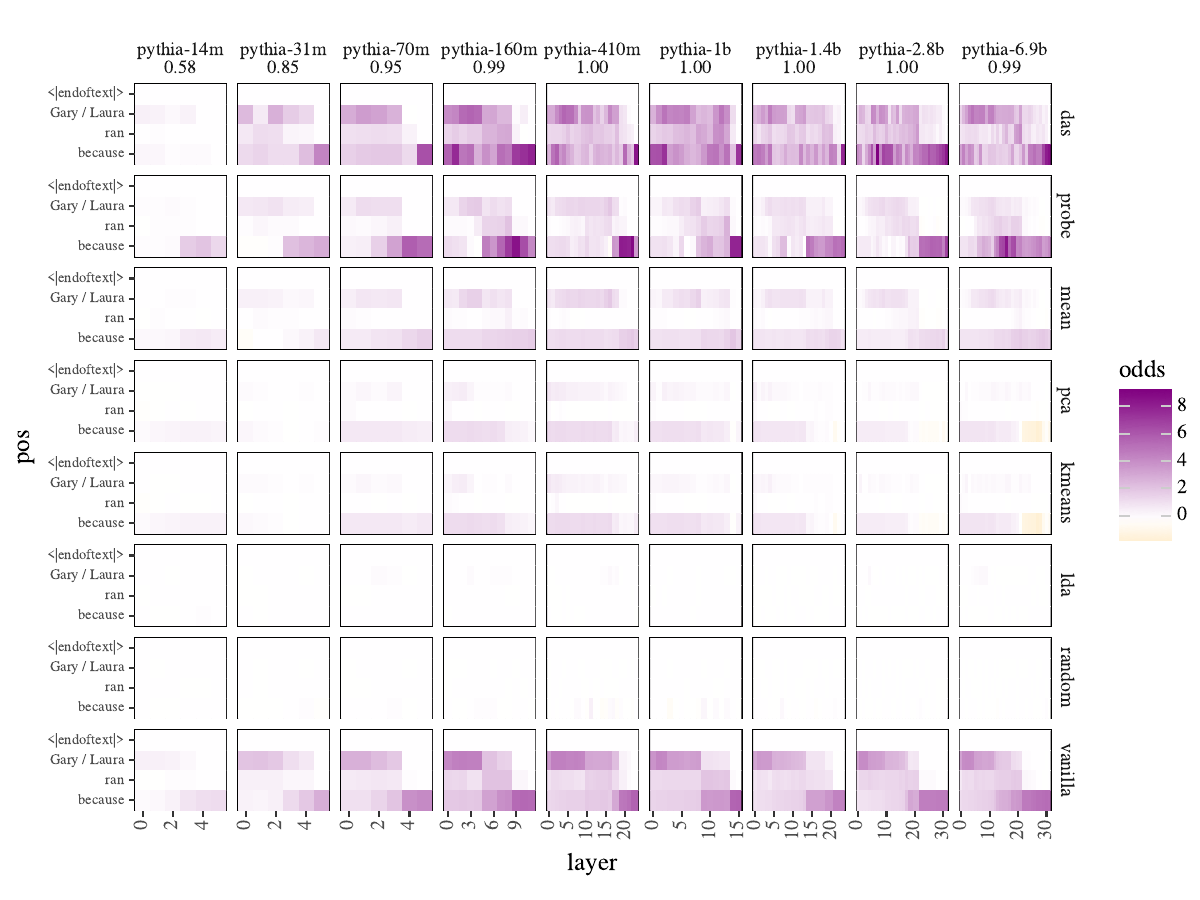}
    \caption{\texttt{agr\_gender}}
\end{figure*}
\begin{figure*}[!hp]
    \centering
    \includegraphics[width=0.85\textwidth]{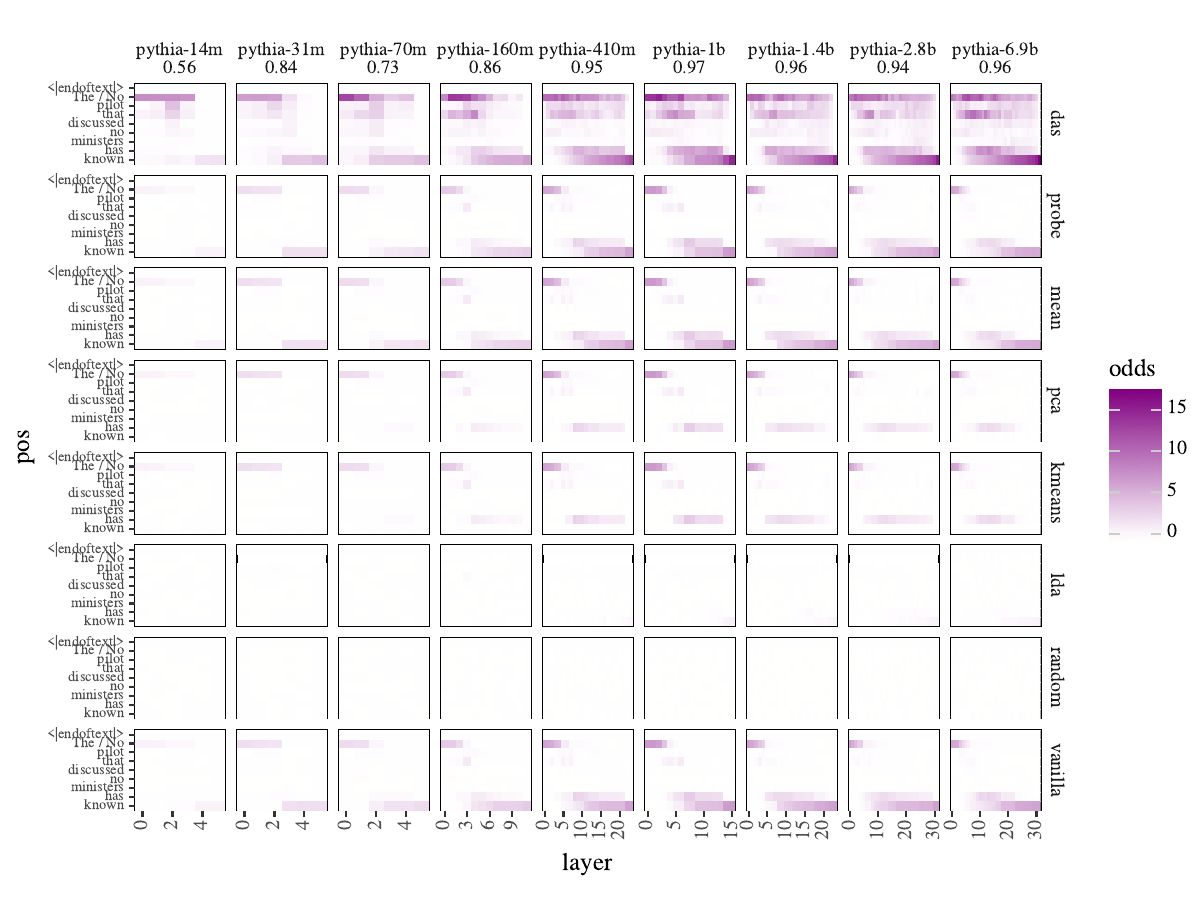}
    \caption{\texttt{npi\_any\_subj-relc}}
\end{figure*}
\begin{figure*}[!hp]
    \centering
    \includegraphics[width=0.85\textwidth]{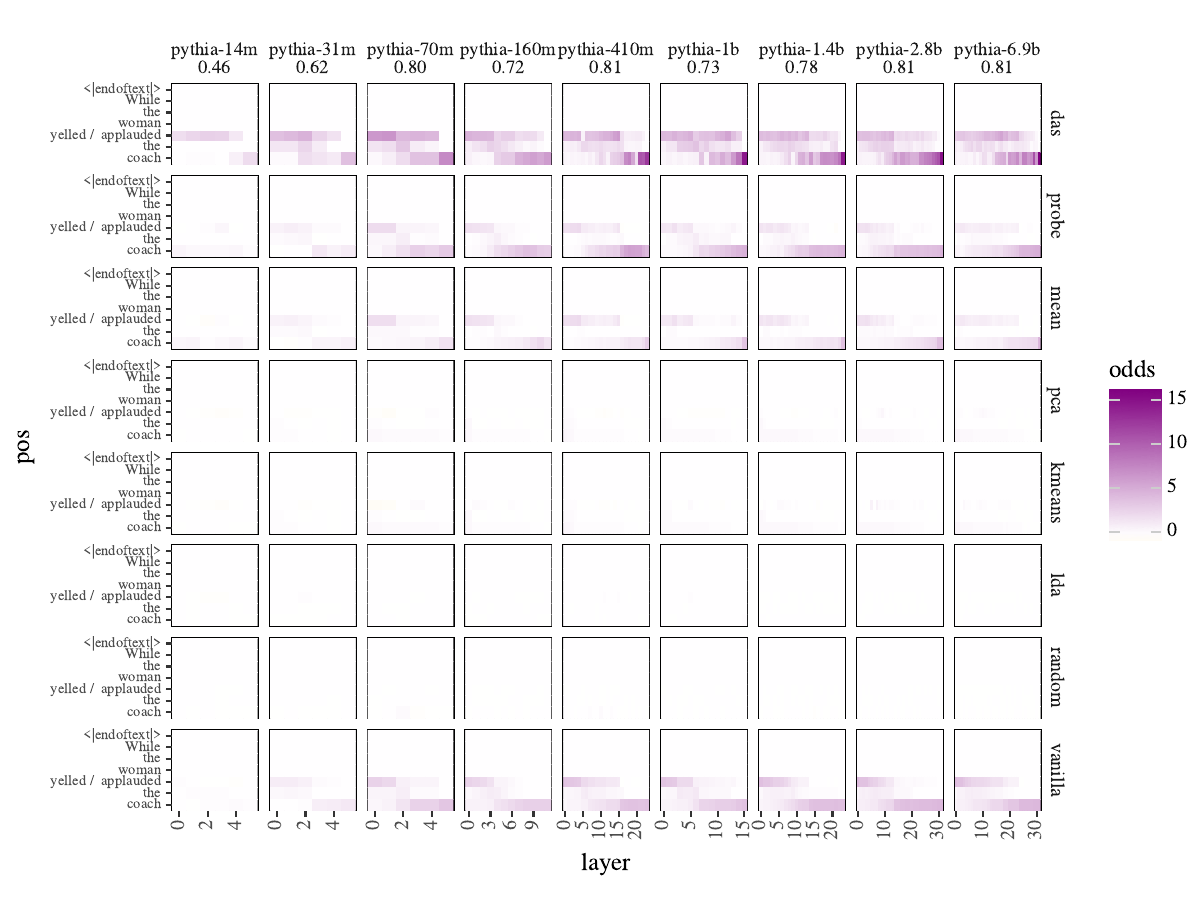}
    \caption{\texttt{garden\_npz\_v-trans}}
\end{figure*}
\begin{figure*}[!hp]
    \centering
    \includegraphics[width=0.85\textwidth]{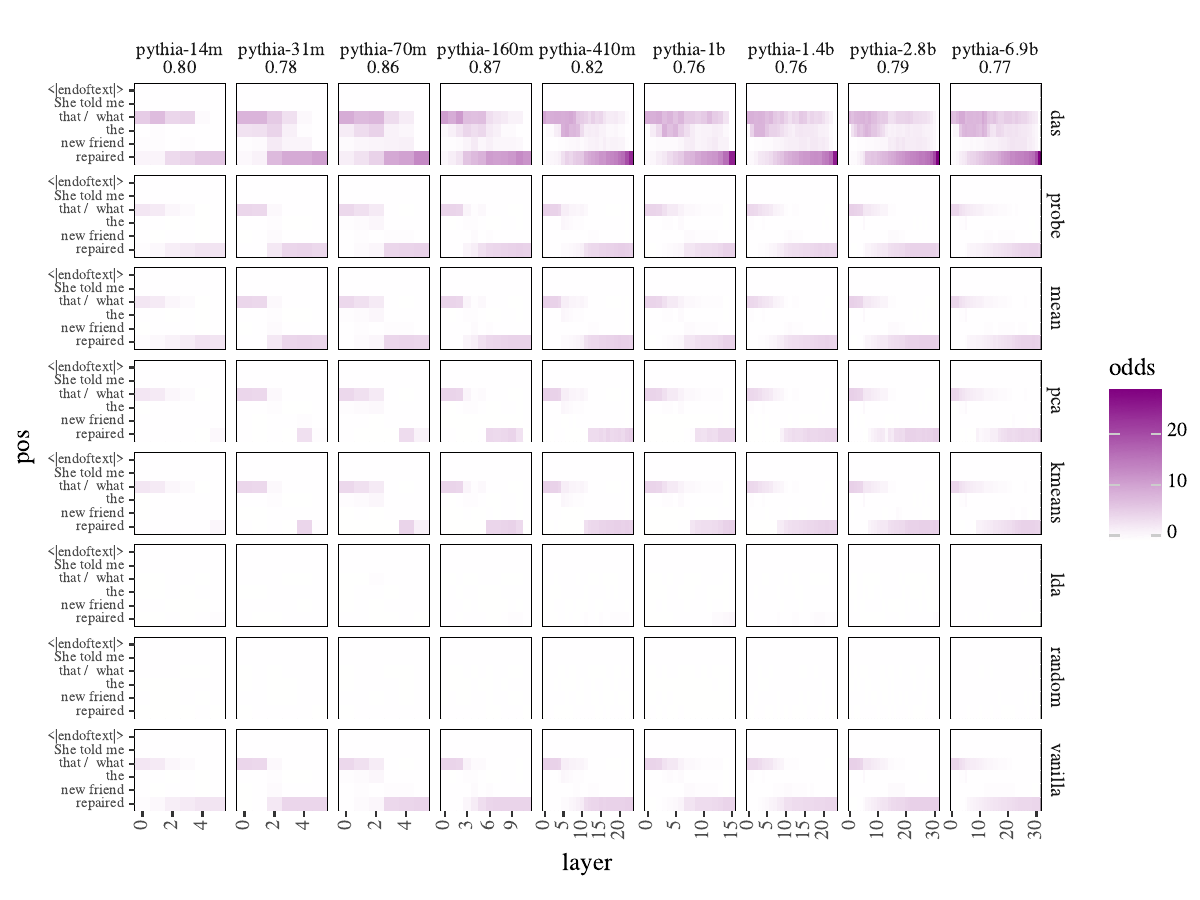}
    \caption{\texttt{filler\_gap\_obj}}
\end{figure*}

\end{document}